\documentclass[letterpaper]{article}
\usepackage{uai2019}
\usepackage[margin=1in]{geometry}
\usepackage{times}

\usepackage{microtype}
\usepackage{times}
\usepackage[usenames,dvipsnames,svgnames,table]{xcolor}
\usepackage{amsthm, amssymb, bbm, bm, mathtools}
\usepackage{enumerate}
\usepackage{graphicx}
\usepackage{float}
\usepackage{wrapfig, subcaption}
\usepackage[font=footnotesize]{caption}
\usepackage[titletoc, toc, page]{appendix}
\usepackage{tabularx}
\usepackage{booktabs,colortbl}
\usepackage{nicefrac}
\usepackage{blindtext}
\usepackage{setspace}
\usepackage{marginnote}
\usepackage{multirow}
\usepackage{csquotes}
\usepackage[scr=boondoxo]{mathalfa}

\usepackage[boxruled, vlined, linesnumbered]{algorithm2e}
\SetAlFnt{\small}
\SetAlCapFnt{\small}
\SetAlCapNameFnt{\small}
\usepackage{algorithmic}
\algsetup{linenosize=\tiny}
\let\oldnl\nl
\newcommand{\nonl}{\renewcommand{\nl}{\let\nl\oldnl}}
\usepackage{etoolbox}
\AtBeginEnvironment{algorithm}{\onehalfspacing}

\usepackage[bookmarks=false,pdfencoding=auto,psdextra]{hyperref}
\hypersetup{
colorlinks=true, linkcolor=Sepia, citecolor=Sepia, filecolor=magenta, urlcolor=NavyBlue,
}
\usepackage{url, bookmark}

\makeatletter
\def\th@plain{%
  \thm@notefont{}
  \itshape 
}
\def\th@definition{%
  \thm@notefont{}
  \normalfont 
}
\makeatother

\usepackage[capitalize,nameinlink]{cleveref}
\crefformat{equation}{(#2#1#3)}
\crefmultiformat{equation}{(#2#1#3)}%
{ and~(#2#1#3)}{, (#2#1#3)}{ and~(#2#1#3)}
\crefrangeformat{section}{Sections~#3#1#4--#5#2#6}
\crefformat{theorem}{Theorem~#2#1#3}
\crefformat{lemma}{Lemma~#2#1#3}
\crefformat{remark}{Remark~#2#1#3}
\crefformat{section}{Section~#2#1#3}
\crefformat{assumption}{Assumption~#2#1#3}
\crefformat{example}{Example~#2#1#3}

\crefrangeformat{equation}{~(#3#1#4--#5#2#6)}
\crefrangeformat{figure}{Figs.~#3#1#4--#5#2#6}

\newcommand{\reals}{\mathbb{R}}

\newcommand{\tbf}[1]{\textbf{#1}}

\DeclarePairedDelimiter\abs{\lvert}{\rvert}
\DeclarePairedDelimiter\norm{\lVert}{\rVert}

\newcommand{\ag}[1]{\ensuremath \left\langle#1\right\rangle}

\DeclareMathOperator*{\argmin}{arg\;min}
\DeclareMathOperator*{\argmax}{arg\;max}

\newcommand{\aeq}[1]{\begin{align} #1 \end{align}}

\newcommand{\aed}[1]{\begin{aligned} #1 \end{aligned}}
\newcommand{\beq}[1]{\begin{equation}#1\end{equation}}

\newcommand{\trm}[1]{\mathrm{#1}}
\newcommand{\enum}[2][(a)]{\begin{enumerate}[#1]{#2}\end{enumerate}}

\newcommand{\la}{\ \leftarrow\ }

\providecommand\f[2]{\ensuremath \frac{#1}{#2}}
\providecommand\rbrac[1]{\ensuremath \left(#1\right)}

\providecommand\RBRAC[1]{\ensuremath \Big(#1 \Big)}
\providecommand\sqbrac[1]{\ensuremath \left[#1\right]}
\providecommand\Sqbrac[1]{\ensuremath \big[#1 \big]}
\providecommand\SQBRAC[1]{\ensuremath \Big[#1 \Big]}

\providecommand\cbrac[1]{\ensuremath \left\{#1\right\}}
\providecommand\Cbrac[1]{\ensuremath \big\{#1 \big\}}
\providecommand\CBRAC[1]{\ensuremath \Big\{#1 \Big\}}

\newcommand{\grad}{\nabla}

\newtheorem{theorem}{Theorem}

\newtheorem{lemma}[theorem]{Lemma}

\theoremstyle{definition}
\newtheorem{definition}[theorem]{Definition}

\newtheorem{remark}[theorem]{Remark}

\let\P\relax
\DeclareMathOperator*{\P}{\mathbb{P}}
\DeclareMathOperator*{\E}{\mathbb{E}}

\definecolor{color_skyblue}{rgb}{0.01,0.39,0.75}

\newcommand{\w}{\omega}
\renewcommand{\r}{\rho}
\renewcommand{\t}{\tau}
\renewcommand{\th}{\theta}
\renewcommand{\a}{\alpha}

\renewcommand{\b}{\beta}
\newcommand{\g}{\gamma}
\renewcommand{\d}{\delta}

\renewcommand{\l}{\lambda}

\def \DD {\mathcal{D}}

\def \AA {\mathcal{A}}

\def \SS {\mathcal{S}}
\def \bb {\mathscr{b}}

\def \vphi {{v_\phi}}

\def \vpi {v^\pi}
\def \qpi {q^\pi}
\def \pith {{\pi_\th}}

\def \qpith {q^\pith}
\def \vpith {v^\pith}
\def \hqpith {\hat{q}^\pith}
\def \hvpith {\hat{v}^\pith}

\def \hapith {\hat{A}^\pith}

\def \dpith {d^\pith}

\def \bb {\mathscr{b}}

\def \pip {{\b}}
\def \dpip {d^{\pip}}

\def \br {\overline{\r}}

\def \ess {\trm{ESS}}

\def \d {\trm{d}}

\def \gon  {\grad^{\trm{on}}}
\def \goff {\grad^{\trm{off}}}

\def \kl {\trm{KL}}

\def \qw {q_\w}

\def \klmax {\max_{s \in \SS}\ \kl}
\usepackage{breqn}

\usepackage{natbib}

\usepackage{titlesec}
\titleformat{\subsection}[block]{\bfseries}{\thesubsection}{0.5em}{\MakeUppercase}

\title{P3O: Policy-on Policy-off Policy Optimization}
\author{
\tbf{Rasool Fakoor}\thanks{
 Correspondence to: Rasool Fakoor [fakoor@amazon.com] and Pratik Chaudhari [prtic@amazon.com].} \\
Amazon Web Services\\
\And
\tbf{Pratik Chaudhari} \\
Amazon Web Services\\
\And
\tbf{Alexander J. Smola} \\
Amazon Web Services\\
}

\newcommand{\ignore}[1]{}

\graphicspath{{./fig_p3o/}}

\begin{document}
\maketitle

\begin{abstract}
On-policy reinforcement learning (RL) algorithms have high sample complexity while off-policy algorithms are difficult to tune. Merging the two holds the promise to develop efficient algorithms that generalize across diverse environments. It is however challenging in practice to find suitable hyper-parameters that govern this trade off. This paper develops a simple algorithm named P3O that interleaves off-policy updates with on-policy updates. P3O uses the effective sample size between the behavior policy and the target policy to control how far they can be from each other and does not introduce any additional hyper-parameters. Extensive experiments on the Atari-2600 and MuJoCo benchmark suites show that this simple technique is effective in reducing the sample complexity of state-of-the-art algorithms. Code to reproduce experiments in this paper is at \href{https://github.com/rasoolfa/P3O}{https://github.com/rasoolfa/P3O}.
\end{abstract}

\section{INTRODUCTION}
\label{s:intro}

Reinforcement Learning (RL) refers to techniques where an agent learns a policy that optimizes a given performance metric from a sequence of interactions with an environment. There are two main types of algorithms in reinforcement learning. In the first type, called on-policy algorithms, the agent draws a batch of data using its current policy. The second type, known as off-policy algorithms, reuse data from old policies to update the current policy.
Off-policy algorithms such as Deep Q-Network~\citep{mnih2015human, mnih2013playing} and Deep Deterministic Policy Gradients DDPG~\citep{lillicrap2015continuous} are biased~\citep{gu2017interpolated} because behavior of past policies may be very different from that of the current policy and hence old data may not be a good candidate to inform updates of the current policy. Therefore, although off-policy algorithms are data efficient, the bias makes them unstable and difficult to tune~\citep{fujimoto2018addressing}. On-policy algorithms do not usually incur a bias~\footnote{Implementations of RL algorithms typically use the undiscounted state distribution instead of discounted distribution, which results in a bias. However, as~\citet{pmlr-v32-thomas14} show, being unbiased is not necessarily good and may even hurt performance.}; they are typically easier to tune~\citep{schulman2017proximal} with the caveat that since they look at each data sample only once, they have poor sample efficiency. Further, they tend to have high variance gradient estimates which necessitates a large number of online samples and highly distributed training~\citep{ilyas2018deep,mnih2016asynchronous}.

Efforts to combine the ease-of-use of on-policy algorithms with the sample efficiency of off-policy algorithms have been fruitful~\citep{gu2016q,o2016pgq,wang2016sample,gu2017interpolated, Nachum2017BridgingTG,degris2012off}. These algorithms merge on-policy and off-policy updates to trade-off the variance of the former against the bias of the latter. Implementing these algorithms in practice is however challenging: RL algorithms already have a lot of hyper-parameters~\citep{henderson2018deep} and such a combination further exacerbates this. This paper seeks to improve the state of affairs.

We introduce the Policy-on Policy-off Policy Optimization (P3O) algorithm in this paper. It performs gradient ascent using the gradient
\beq{
    \aed{
        &\E_{\substack{s \sim \dpith,\\ a\sim \pith}} \SQBRAC{\grad \log \pith \hapith} + \E_{\substack{s \sim \dpip,\\ a\sim \pip}} \SQBRAC{\min(\r, c)  \hapith \grad \log \pith}\\
        &\quad \qquad \qquad -\l \grad_\th \E_{s \sim \dpip, a \sim \pip} \kl \RBRAC{\pip(\cdot|s)\ ||\ \pith(\cdot|s)}
    }
    \label{eq:p3o_main}
}
where the first term is the on-policy policy gradient, the second term is the off-policy policy gradient corrected by an importance sampling (IS) ratio $\r$ and the third term is a constraint that keeps the state distribution of the target policy $\pith$ close to that of the behavior policy $\pip$. Our key contributions are:
\enum[1.]
{
    \item we automatically tune the IS clipping threshold $c$ and the $\kl$ regularization coefficient $\l$ using the normalized effective sample size (ESS), and
    \item we control changes to the target policy using samples from replay buffer via an explicit Kullback-Leibler constraint.
}
The normalized ESS measures how efficient off-policy data is to estimate the on-policy gradient. We set
\[
    \l = 1-\ess \quad \trm{and} \quad c = \ess.
\]
We show in~\cref{s:expt} that this simple technique leads to consistently improved performance over competitive baselines on discrete action tasks from the Atari-2600 benchmark suite~\citep{bellemare2013arcade} and continuous action tasks from MuJoCo benchmark~\citep{todorov2012mujoco}.

\section{BACKGROUND}
\label{s:background}

Consider a discrete-time agent that interacts with the environment. The agenet picks an action $a \in \AA$ given the current state $s \in \SS$ using a policy $\pi(a | s)$. It receives a reward $r(s, a) \in \reals$ after this interaction and its objective is to maximize the discounted sum of rewards $G_t = \sum_{i=t}^\infty\ \g^{i-t}\ r(s_i, a_i)$ where $\g \in [0,1)$ is a scalar constant that discounts future rewards. The quantity $G_t$ is called the return. We shorten $r(s_t, a_t)$ to $r_t$ to simplify notation.

If the initial state $s_0$ is drawn from a distribution $d^0(s)$ and the agent follows the policy $\pi$ thereafter, the action-value function and the state-only value function are
\beq{
    \aed{
        \qpi(s_t, a_t) &= \E_{(s,a) \sim \pi} \SQBRAC{G_t | s_t, a_t},\quad \textrm{and}\\
        \vpi(s_t) &= \E_{a_t} \SQBRAC{\qpi(s_t, a_t)}
    }
    \label{eq:qpi_vpi}
}
respectively. The best policy $\pi^* = \argmax_\pi J(\pi)$ maximizes the expected value of the returns where
\beq{
    J(\pi) = \E_{s \sim d^0} \SQBRAC{\vpi(s)}.
    \label{eq:jpi}
}

\subsection{Policy Gradients}
\label{ss:pg}

We denote by $\pith$, a policy that is parameterized by parameters $\th \in \reals^n$. This induces a parameterization of the state-action and state-only value functions which we denote by $\qpith$ and $\vpith$ respectively. Monte-Carlo policy gradient methods such as REINFORCE~\citep{Williams1992} solve for the best policy $\pith^*$, typically using first-order optimization, using the likelihood-ratio trick to compute the gradient of the objective. Such a policy gradient of~\cref{eq:jpi} is given by
\beq{
    \E_{s_t \sim \dpith,\ a_t \sim \pith} \SQBRAC{\qpith(s_t, a_t)\ \grad_\th \log \pith(a_t | s_t)}
    \label{eq:pg_vanilla}
}
where $\dpith$ is the unnormalized discounted state visitation frequency $\dpith(s) = \sum_{t=0}^\infty \g^t \P(s_t = s)$.

\begin{remark}[Variance reduction]
\label{rem:pg_baseline}
The integrand in~\cref{eq:pg_vanilla} is estimated in a Monte-Carlo fashion using sample trajectories drawn using the current policy $\pith$. The action-value function $\qpith$ is typically replaced by $\hqpith(s_t,a_t) = \sum_{i=0}^\infty \g^i r_{t+i}$. Both of these approximations entail a large variance for policy gradients~\citep{kakade2002approximately,baxter2001infinite} and a number of techniques exist to mitigate the variance. The most common one is to subtract a state-dependent control variate (baseline) $\hvpith(s)$ from $\hqpith(s)$. This leads to the Monte-Carlo estimate of the advantage function~\citep{Konda2000}
\[
    \hapith(s,a) = \hqpith(s,a) - \hvpith(s)
\]
which is used in place of $\qpith$ in~\cref{eq:pg_vanilla}. Let us note that more general state-action dependent baselines can also be used~\citep{liu2017action}. We denote the baselined policy gradient integrand in short by $g(\pith) = \hapith(s, a)\ \grad_\th \log \pith(a | s)$ to rewrite~\cref{eq:pg_vanilla} as
\beq{
    \gon_\th J(\pith) = \E_{s \sim \dpith,\ a\sim \pith}\ \SQBRAC{g(\pith)}.
    \label{eq:pg}
}
\end{remark}

\subsection{Off-policy policy gradient}
\label{ss:off_policy_pg}

The expression in~\cref{eq:pg} is an expectation over data collected from the current policy $\pith$. Vanilla policy gradient methods use each datum only once to update the policy which makes then sample inefficient. A solution to this problem is to use an experience replay buffer~\citep{lin1992self} to store previous data and reuse these experiences to update the current policy using importance sampling. For a mini-batch of size $T$ consisting of $\cbrac{(s_k,a_k,s_k')}$ with $k \leq T$, the integrand in~\cref{eq:pg} becomes
\[
    \RBRAC{\prod_{t=0}^T \r(s_t, a_t)} \sum_{t=0}^T \rbrac{\sum_{i=0}^{T-t} \g^i r_{t+i} } \grad_\th \log \pith(a_t | s_t)
\]
where the importance sampling (IS) ratio
\beq{
    \r(s,a) = \f{\pith(a | s)}{\pip(a | s)} > 0
    \label{eq:is}
}
governs the relative probability of the candidate policy $\pith$ with respect to $\pip$.

\citet{degris2012off} employed marginal value functions to approximate the above gradient and they obtained the expression
\beq{
    \E_{s \sim \dpip, a \sim \pip} \SQBRAC{\r(s,a)\  \hapith(s,a)\ \grad_\th \log \pith(a | s)}
    \label{eq:pg_off_no_clip}
}
for the off-policy policy gradient. Note that states are sampled from $\dpip$ which is the discounted state distribution of $\pip$. Further, the expectation occurs using the policy $\pip$ while the action-value function $\qpith$ is that of the target policy $\pith$. This is important because in order to use the off-policy policy gradient above, one still needs to estimate $\qpith$. The authors in~\citet{wang2016sample} estimate $\qpith$ using the Retrace($\l$) estimator~\citep{munos2016safe}. If $\pith$ and $\pip$ are very different from each other (i) the importance ratio $\r(s,a)$ may vary across a large magnitude, and (ii) the estimate of $\qpith$ may be erroneous. This leads to difficulties in estimating the off-policy policy gradient in practice. An effective way to mitigate (i) is to clip $\r(s,a)$ at some threshold $c$. We will use this clipped importance ratio often and denote it as $\br_c = \min(\r, c)$. This helps us shorten the notation for the off-policy policy gradient to
\beq{
    \goff_\th J(\pith) = \E_{s \sim \dpip, a \sim \pip} \SQBRAC{\br_c\ g(\pith)}.
    \label{eq:pg_off}
}

\subsection{Covariate Shift}
\label{ss:covariate_shift}

Consider the supervised learning where we observe iid data from a distribution $q(x)$, say the training dataset. We would however like to minimize the loss on data from another distribution $p(x)$, say the test data. This amounts to minimizing
\beq{
    \aed{
        \E_{x \sim p(x)}\ &\E_{y|x}\ \sqbrac{\ell(y, \varphi(x))}\\
        &= \E_{x\sim q(x)}\ \E_{y|x}\ \sqbrac{w(x)\ \ell(y, \varphi(x))}.
    }
    \label{eq:is}
}
Here $y$ are the labels associated to draws $x \sim q(x)$ and $\ell(y, \varphi(x))$ is the loss of the predictor $\varphi(x)$. The importance ratio is
\beq{
    w(x) := \f{\d p(x)}{\d q(x)}
}
is the Radon-Nikodym derivative of the two densities~\citep{resnick2013probability} and it re-balances the data to put more weight on unlikely samples in $q(x)$ that are likely under the test data $p(x)$. If the two distributions are the same, the importance ratio is 1 and this is unnecessary. When the two distributions are not the same, we have an instance of covariate shift and need to use the trick in~\cref{eq:is}.

\begin{definition}[Effective sample size]
\label{def:ess}
Given a dataset $X = \cbrac{x_1, x_2, \ldots, x_N}$ and two densities $p(x)$ and $q(x)$ with $p(x)$ being absolutely continuous with respect to $q(x)$, the effective sample size is defined as the number of samples from $p(x)$ that would provide an estimator with a performance equal to that of the importance sampling (IS) estimator in~\cref{eq:is} with $N$ samples~\citep{kong1992}. For our purposes, we will use the normalized effective sample size
\beq{
    \ess = \f{1}{N}\ \norm{w(X)}_1^2/\norm{w(X)}_2^2
    \label{eq:ess}
}
where $w(X) := [\d p(x_1)/\d q(x_1), \ldots, \d p(x_N)/\d q(x_N)]$ is a vector that consists of  evaluated at the samples. This expression is a good rule of thumb and is occurs, for instance, for a weighted average of Gaussian random variables~\citep{quionero2009dataset} or in particle filtering~\citep{smith2013sequential}. We have normalized the ESS by the size of the dataset which makes $\ess \in [0,1]$.
\end{definition}

Note that estimating the importance ratio $w(x)$ requires the knowledge of both $p(x)$ and $q(x)$. While this is not usually the case in machine learning, reinforcement learning allows us access to both off-policy data and the on-policy data easily. We can therefore estimate $w(x)$ easily in RL. We can use the ESS as an indicator of the efficacy of updates to $\pith$ with samples drawn from the behavior policy $\pip$. If the ESS is large, the two policies predict similar actions given the state and we can confidently use data from $\pip$ to update $\pith$.


\section{APPROACH}
\label{s:approach}

This section discusses the P3O algorithm. We first identify key characteristics of merging off-policy and on-policy updates and then discuss the details of the algorithm and provide insight into its behavior using ablation experiments.

\subsection{Combining on-policy and off-policy gradients}
\label{ss:combining_on_off_pg}

We can combine the on-policy update~\cref{eq:pg} with the off-policy update~\cref{eq:pg_off} after bias-correction on the former as
\beq{
    \E_{\substack{s\sim \dpith,\\ a\sim \pith}} \SQBRAC{\rbrac{1 - \f{c}{\r}}_+ g(\pith)} + \E_{\substack{s \sim \dpip,\\ a \sim \pip}} \SQBRAC{\br_c\ g(\pith)},
    \label{eq:pg_acer}
}
where $\rbrac{\cdot}_+ := \max\rbrac{\cdot, 0}$. This is similar to the off-policy actor-critic~\citep{degris2012off} and ACER gradient~\citep{wang2016sample} except that the authors in~\citet{wang2016sample} use the Retrace($\l$) estimator to estimate $\qpith$ in~\cref{eq:pg_off}. The expectation in the second term is computed over actions that were sampled by $\pip$ whereas the expectation of the first term is computed over all actions $a \in \AA$ weighted by the probability of taking them $\pith(a | s)$. The clipping constant $c$ in~\cref{eq:pg_acer} controls the off-policy updates versus on-policy updates. As $c \to \infty$, ACER does a completely off-policy update while we have a completely on-policy update as $c \to 0$. In practice, it is difficult to pick a value for $c$ that works well for different environments as we elaborate upon in the following remark. This difficulty in choosing $c$ is a major motivation for the present paper.

\begin{figure}[!htpb]
\centering
\includegraphics[width=0.3\textwidth]{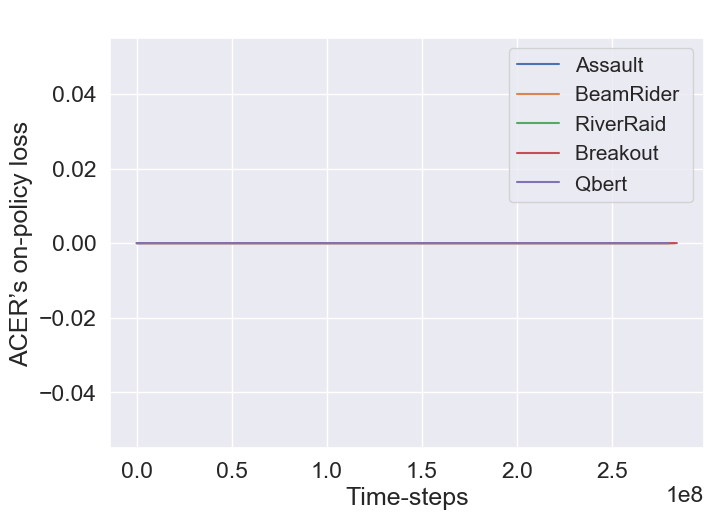}
\caption{\tbf{On-policy loss for ACER is zero all through training due to aggressive importance ratio thresholding.} ACER had the highest reward from among A2C, PPO and P3O in 3 out of these 5 games (Assault, RiverRaid and BreakOut; see the Supplementary Material for more details). In spite of the on-policy loss being zero for all Atari games, ACER receives good rewards across the benchmark.}
\label{fig:acer_loss}
\end{figure}

\begin{remark}[How much on-policy updates does ACER do?]
\label{rem:acer_on_policy}
We would like to study the fraction of weight updates coming from on-policy data as compared to those coming from off-policy data in~\cref{eq:pg_acer}. We took a standard implementation of ACER\footnote{OpenAI baselines: \href{https://github.com/openai/baselines}{https://github.com/openai/baselines}} with published hyper-parameters from the original authors ($c=10$) and plot the on-policy part of the loss (first term in~\cref{eq:pg_acer}) as training progresses in~\cref{fig:acer_loss}. The on-policy loss is zero throughout training. This suggests that the performance of ACER~\citep{wang2016sample} should be attributed pre-dominantly to off-policy updates and the Retrace($\l$) estimator rather than the combination of off-policy and on-policy updates. This experiment demonstrates the importance of hyper-parameters when combining off-policy and on-policy updates, it is difficult tune hyper-parameters that combine the two and work in practice.
\end{remark}

\subsection{Combining on-policy and off-policy data with control variates}
\label{ss:ipg}
Another way to leverage off-policy data is to use it to learn a control variate, typically the action-value function $\qw$. This has been the subject of a number papers; recent ones include Q-Prop~\citep{gu2016q} which combines Bellman updates with policy gradients and Interpolated Policy Gradients (IPG)~\citep{gu2017interpolated} which directly interpolates between on-policy and off-policy deterministic gradient, DPG and DDPG algorithms,~\citep{silver2014deterministic,Lillicrap2016ContinuousCW}) using a hyper-parameter. To contrast with the ACER gradient in~\cref{eq:pg_acer}, the IPG is
\beq{
    (1-\nu) \E_{\substack{s \sim \dpith,\\a \sim \pith}} \SQBRAC{g(\pith)} + \nu\ \grad_\th \E_{\substack{s \sim \dpip,\\ a \sim \pith(a | s)}} \SQBRAC{\qw(s, a)}
    \label{eq:ipg}
}
where $q_w$ is an off-policy fitted critic. Notice that since the policy $\pith$ is stochastic the above expression uses $\grad_\th \E_{a \sim \pith} \Cbrac{\qw}$ for the off-policy part instead of the DPG $\grad_\th \qw(s, \mu_\th(s))$ for a deterministic policy $\mu_\th(s)$. This avoids training a separate deterministic policy (unlike Q-Prop) for the off-policy part and encourages on-policy exploration and an implicit trust region update. The parameter $\nu$ explicitly controls the trade-off between the bias and the variance of off-policy and on-policy gradients respectively. However, we have found that it is difficult to pick this parameter in practice; this is also seen in the results of~\citep{gu2017interpolated} which show sub-par performance on MuJoCo~\citep{todorov2012mujoco} benchmarks; for instance compare these results to similar experiments in~\cite{fujimoto2018addressing} for the Twin Delayed DDPG (TD3) algorithm.

\subsection{P3O: Policy-on Policy-off Policy optimization}
\label{ss:p3o}

Our proposed approach, named Policy-on Policy-off Policy Optimization (P3O) explicitly controls the deviation of the target policy from the behavior policy. The gradient of P3O is given by
\beq{
    \aed{
        \E_{s\sim \dpith, a\sim \pith} &\SQBRAC{g(\pith)} + \E_{s \sim \dpip, a\sim \pip} \SQBRAC{\br_c\ g(\pith)}\\
        &- \l \grad_\th \E_{s \sim \dpip, a\sim \pip} \kl \RBRAC{\pip(\cdot|s)\ ||\ \pith(\cdot|s)}.
    }
    \label{eq:p3o}
}
The first term above is the standard on-policy gradient. The second term is the off-policy policy gradient with truncation of the IS ratio using a constant $c$ while the third term allows explicit control of the deviation of the target policy $\pith$ from $\pip$. We do not perform bias correction in the first term so it is missing the factor $\rbrac{1 -\f{c}{\r}}_+$ from the ACER gradient~\cref{eq:pg_acer}. As we noted in~\cref{rem:acer_on_policy}, it may be difficult to pick a value of $c$ which keeps this factor non-zero. Even if the $\kl$-term is zero, the above gradient is a biased estimate of the on-policy policy gradient. Further, the $\kl$-divergence term can be rewritten as $\E_{s\sim \dpip, a\sim \pip} \sqbrac{\log \r}$ and therefore minimizes the importance ratio $\r$ over the entire replay buffer $\pip$. There are two hyper-parameters in the P3O gradient: the IS ratio threshold $c$ and the $\kl$ regularization co-efficient $\l$. We use the following reasoning to pick them.

If the behavior and target policies are far from each other, we would like the $\l$ be large so as to push them closer. If they are too similar to each other, it entails that we could have performed more exploration, in this scenario, we desire a smaller regularization co-efficient $\l$. We set
\beq{
    \l = 1 - \ess
    \label{eq:l_ess}
}
where the ESS in~\cref{eq:ess} is computed using the current mini-batch sampled from the replay buffer $\pip$.

The truncation threshold $c$ is chosen to keep the variance of the second term small. Smaller the $c$, less efficient the off-policy update and larger the $c$ higher the variance of this update. We set
\beq{
    c = \ess.
    \label{eq:c_ess}
}
This is a very natural way to threshold the IS factor $\r(s,a)$ because $\ess \in [0,1]$. This ensures an adaptive trade-off between the reduced variance of the gradient estimate and the inefficiency of a small IS ratio $\r$. Note that the ESS is computed on a mini-batch of transitions and their respective IS factors and hence clipping an individual $\r(s,a)$ using the ESS tunes $c$ automatically to the mini-batch.

The gradient of P3O in~\cref{eq:p3o} is motivated by the following observation: explicitly controlling the $\kl$-divergence between the target and the behavior policy encourages them to have the same visitation frequencies. This is elaborated upon by~\cref{lem:gap_state_distribution} which follows from the time-dependent state distribution bound proved in~\citep{schulman2015trust,kahn2017plato}.

\begin{figure*}[!htpb]
\centering
\captionsetup[subfigure]{justification=centering}
     \begin{subfigure}[t]{0.35\textwidth}
         \centering
         \includegraphics[width=\textwidth]{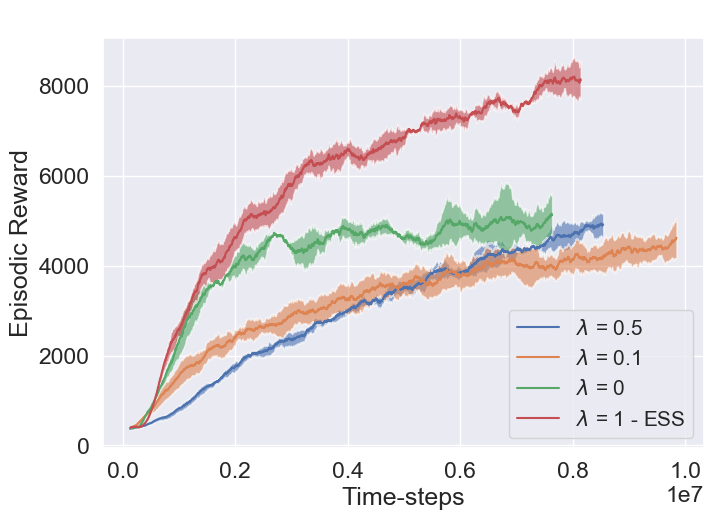}
         \caption{BeamRider}
         \label{fig:effect_of_lambda:beam_rider}
     \end{subfigure}
     \hspace{0.1in}
     \begin{subfigure}[t]{0.35\textwidth}
         \centering
         \includegraphics[width=\textwidth]{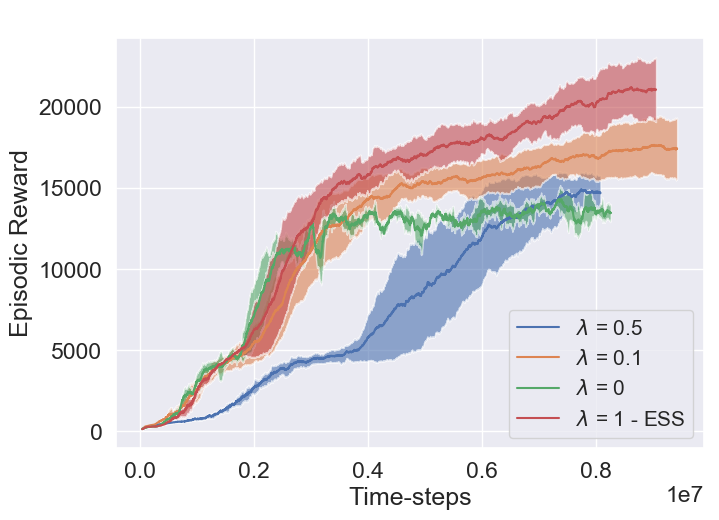}
         \caption{Qbert}
         \label{fig:effect_of_lambda:qbert}
     \end{subfigure}
\caption{\tbf{Effect of $\l$ on performance.} First, a non-zero value of $\l$ trains much faster than without the $\kl$ regularization term because the target policy is constrained to be close to an entropic $\pip$. Second, for hard exploration games like Qbert, a smaller value $\l=0.1$ works much better than $\l=0.5$ while the trend is somewhat reversed for easy exploration games such as BeamRider. The ideal value of $\l$ thus depends on the environment and is difficult to pick before-hand. Setting $\l=1-\ess$ tunes the regularization adaptively depending upon the particular mini-batch and works significantly better for easy exploration, it also leads to gains in hard exploration tasks.}
\label{fig:effect_of_lambda}
\end{figure*}

\begin{lemma}[Gap in discounted state distributions]
\label{lem:gap_state_distribution}
The gap between the discounted state distributions $\dpith$ and $\dpip$ is bounded as
\beq{
    \norm{\dpith - \dpip}_1 \leq \f{2 \g}{(1-\g)^2}\ \sqrt{\klmax(\pip\ ||\ \pith)}
    \label{eq:diff_dpith}
}
\end{lemma}
The $\kl$-divergence penalty in~\cref{eq:p3o} is directly motivated from the above lemma; we however use $\E_{s \sim \dpip, a\sim \pip} \Sqbrac{\kl(\pith\ ||\ \pip)}$ which is easier to estimate.

\begin{remark}[Effect of $\l$]
\label{rem:effect_of_lambda}
\cref{fig:effect_of_lambda} shows the effect of picking a good value for $\l$ on the training performance. We picked two games in Atari for this experiment: BeamRider which is an easy exploration task and Qbert which is a hard exploration task~\citep{BellemareNIPS2016}. As the figure and the adjoining caption shows, picking the correct value of $\l$ is critical to achieving good sample complexity. The ideal $\l$ also changes as the training progress because policies are highly entropic at initialization which makes exploration easier. It is difficult to tune $\l$ using annealing schedules, this has also been mentioned by the authors in~\citet{schulman2017proximal} in a similar context. Our choice of $\l = 1 -\ess$ adapts the level of regularization automatically.
\end{remark}

\begin{remark}[P3O adapts the bias in policy gradients]
\label{rem:adaptive_bias}
There are two sources of bias in the P3O gradient. First, we do not perform correction of the on-policy term in~\cref{eq:pg_acer}. Second, the KL term further modifies the descent direction by averaging the target policy's entropy over the replay buffer. If $\r(s,a) > c$ for all transitions in the replay buffer, the bias in the P3O update is
\aeq{
    &\E_{\substack{s \sim\dpith,\\ a \sim \pith}}\ \SQBRAC{-\f{c}{\r}  \hapith \grad \log \pith} + \E_{s \sim \dpip, a \in \AA} \SQBRAC{\l \grad \log \pith(a| s)} \notag\\
    &=\E_{s \sim \dpip, a \sim \pip}\ \SQBRAC{-\ess\  \hapith \grad \log \pith} \notag\\
    &\quad \qquad+ \E_{s \sim \dpip, a \in \AA} \SQBRAC{(1-\ess)\ \grad \log \pith(a| s)}
    \label{eq:p3o_bias}
}
The above expression suggests a very useful feature. If the $\ess$ is close to $1$, i.e., if the target policy is close to the behavior policy, P3O is a heavily biased gradient with no entropic regularization. On the other hand, if the ESS is zero, the entire expression above evaluates to zero. The choice $c = \ess$ therefore tunes the bias in the P3O updates adaptively. Roughly speaking, if the target policy is close to the behavior policy, the algorithm is confident and moves on even with a large bias. It is difficult to control the bias coming from the behavior policy, the ESS allows us to do so naturally.

A number of implementations of RL algorithms such as Q-Prop and IPG often have subtle, unintentional biases~\citep{tucker2018mirage}. However, the improved performance of these algorithms, as also that of P3O, suggests that biased policy gradients might be a fruitful direction for further investigation.
\end{remark}

\begin{figure*}[!htpb]
\centering
\captionsetup[subfigure]{justification=centering}
     \begin{subfigure}[t]{0.319\textwidth}
         \centering
         \includegraphics[width=\textwidth]{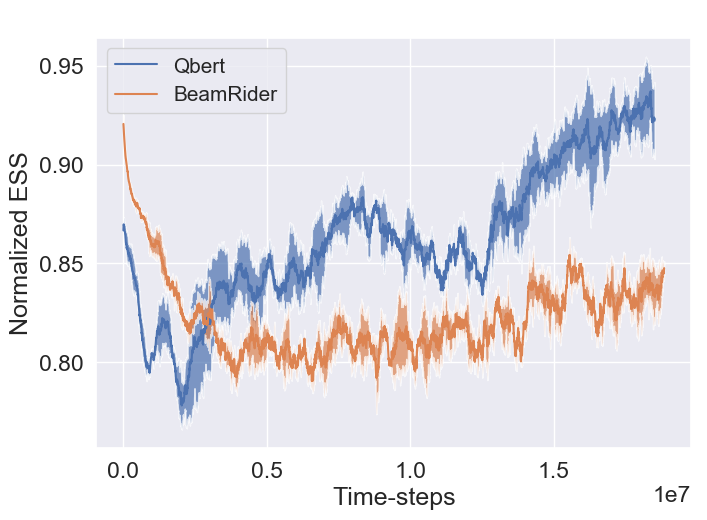}
         \caption{ESS}
         \label{fig:evolution_of_ess:ess}
     \end{subfigure}
     \begin{subfigure}[t]{0.325\textwidth}
         \centering
         \includegraphics[width=\textwidth]{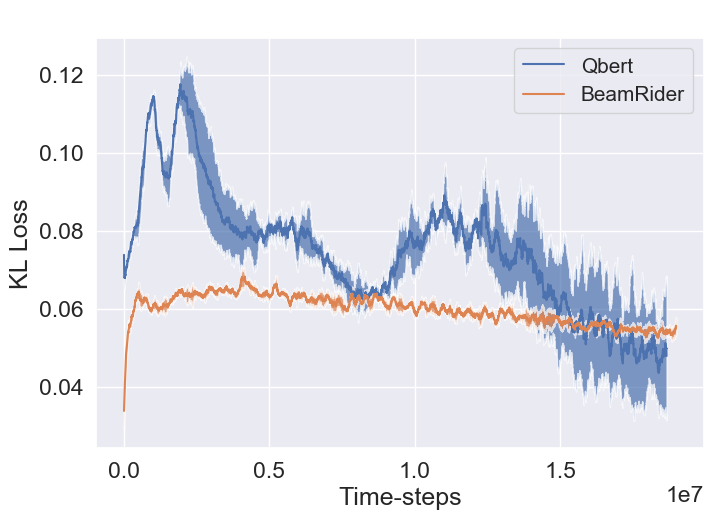}
         \caption{KL term $\E_{s \sim \pip} \SQBRAC{\kl(\pip(\cdot |s)\ ||\ \pith(\cdot|s))}$}
         \label{fig:kl_loss}
     \end{subfigure}
     \begin{subfigure}[t]{0.325\textwidth}
         \centering
         \includegraphics[width=\textwidth]{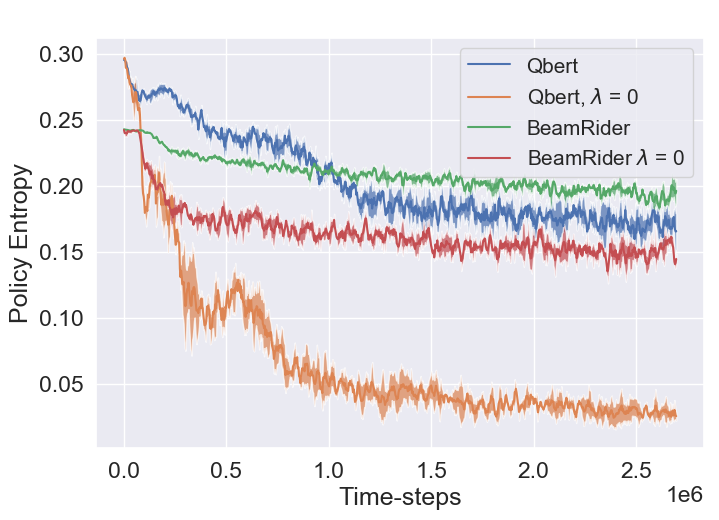}
         \caption{Entropy of $\pith$}
         \label{fig:evolution_of_ess:ent}
     \end{subfigure}
\caption{\tbf{Evolution of ESS, KL penalty and the entropy of $\pith$ as training progresses.} \cref{fig:evolution_of_ess:ess} shows the evolution of normalized ESS. A large value of ESS indicates that the target policy $\pith$ is close to $\pip$ in its state distribution. The ESS is about $0.85$ for a large fraction of the training which suggests a good trade-off between exploration and exploitation. The KL term in~\cref{fig:kl_loss} is relatively constant during the course of training because its coefficient $\l$ is adapted by ESS. This enables the target policy to be exploratory while still being able to leverage off-policy data from the behavior policy. \cref{fig:evolution_of_ess:ent} shows the evolution of the entropy of $\pith$ normalized by the number of actions $\abs{\AA}$. Note that using $\l=0$ results in the target policy having a smaller entropy than standard P3O. This reduces its exploratory behavior and the latter indeed achieves a higher reward as seen in~\cref{fig:effect_of_lambda}.}
\label{fig:evolution_of_ess}
\end{figure*}

\begin{figure*}[!htpb]
\centering
\captionsetup[subfigure]{justification=centering}
     \begin{subfigure}[t]{0.35\textwidth}
         \centering
         \includegraphics[width=\textwidth]{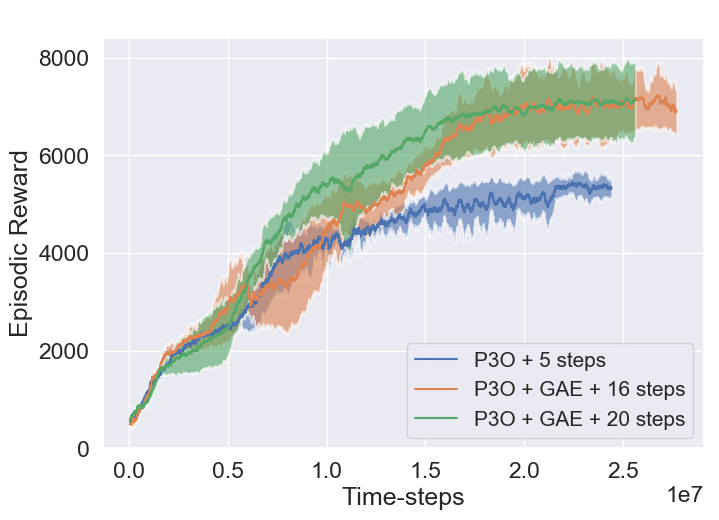}
         \caption{Ms. Pac-Man}
         \label{fig:gae:mspacman}
     \end{subfigure}
     \hspace{0.1in}
     \begin{subfigure}[t]{0.35\textwidth}
         \centering
         \includegraphics[width=\textwidth]{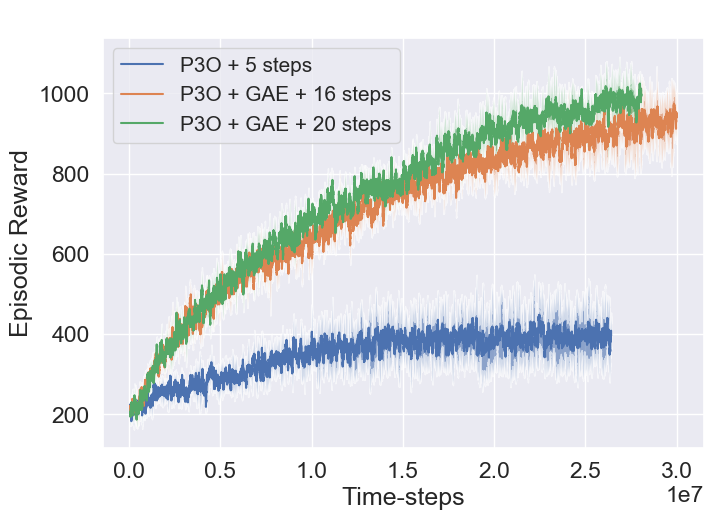}
         \caption{Gravitar}
         \label{fig:gae:gravitar}
     \end{subfigure}
\caption{\tbf{Effect of roll-out length and GAE.} \cref{fig:gae:mspacman,fig:gae:gravitar} show the progress of P3O with and without generalized advantage estimation. GAE leads to significant improvements in performance. The above figures also show the effect of changing the number of time-steps from the environment used in on-policy updates: longer time-horizons help in games with sparse rewards although the benefit diminishes across the suite after $20$ steps.}
\label{fig:gae}
\end{figure*}

\subsection{Discussion on the KL penalty}
\label{ss:trust_region}

The $\kl$-divergence penalty in P3O is reminiscent of trust-region methods. These are a popular way of making monotonic improvements to the policy and avoiding premature moves, e.g., see the TRPO algorithm by~\citet{schulman2015trust}. The theory in TRPO suggests optimizing a surrogate objective where the hard $\kl$ divergence constraint is replaced by a penalty in the objective. In our setting, this amounts to the penalty $\l\ \E_{s \sim \pip} \SQBRAC{\kl(\pip\ || \pith)}$. Note that the behavior policy $\b$ is a mixture of previous policies and this therefore amounts to a penalty that keeps $\pith$ close to \emph{all} policies in the replay buffer $\pip$. This is also done by the authors in~\citet{wang2016sample} to stabilize the high variance of actor-critic methods.

A penalty with respect to \emph{all} past policies slows down optimization. This can be seen abstractly as follows. For an optimization problem $x^* = \argmin_x f(x)$, the gradient update $x^{k+1} = x^k - \a^k \grad f(x^k)$ can be written as
\[
    x^{k+1} = \argmin_y \CBRAC{ \ag{\grad f(x), y} + \f{1}{2\a^k} \norm{y-x^k}^2}
\]
if the $\argmin$ is unique; here $x^k$ is the iterate and $\a^k$ is the step-size at the $k^{\trm{th}}$ iteration. A penalty with respect to all previous iterates $\cbrac{x^1, x^2, \ldots, x^k}$ can be modeled as
\beq{
    x^{k+1} = \argmin_y \CBRAC{ \ag{\grad f(x), y} + \f{1}{2\a^k} \sum_{i=1}^k \norm{y-x^i}^2}
    \label{eq:prox_average}
}
which leads to the update equation
\[
    x^{k+1} = \f{1}{k} \sum_{i=1}^k x^i - \f{\a^k}{k} \grad f(x^k)
\]
which has a vanishing step-size as $k \to \infty$ if the schedule $\a^k$ is left unchanged. We would expect such a vanishing step-size of the policy updates to hurt performance.

The above observation is at odds with the performance of both ACER and P3O; see~\cref{s:expt} which shows that both algorithms perform strongly on the Atari benchmark suite. However~\cref{fig:evolution_of_ess} helps reconcile this issue. As the target policy $\pith$ is trained, the entropy of the policy decreases, while older policies in the replay buffer are highly entropic and have more exploratory power. A penalty that keeps $\pith$ close to $\pip$ encourages $\pith$ to explore. This exploration compensates for the decreased magnitude of the on-policy policy gradient seen in~\cref{eq:prox_average}.

\subsection{Algorithmic details}
\label{ss:algorithmic_details}

The pseudo-code for P3O is given in~\cref{alg:p3o}. At each iteration, it rolls out $K=16$ trajectories of $T=16$ time-steps each using the current policy and appends them to the replay buffer $\DD$. In order to be able to compute the $\kl$-divergence term, we store the policy $\pith(\cdot | s)$ in addition to the action for all states.

P3O performs sequential updates on the on-policy data and the off-policy data. In particular, Line 5 in~\cref{alg:p3o} samples a Poisson random variable that governs the number of off-policy updates for each on-policy update in P3O. This is also commonly done in the literature~\citep{wang2016sample}. We use Generalized Advantage Estimation (GAE)~\citep{schulman2015high} to estimate the advantage function in P3O. We have noticed significantly improved results with GAE as compared to without it, as~\cref{fig:gae} shows.

{
\IncMargin{0.2in}
\SetKwInOut{Input}{Input}
\begin{algorithm}
\small
\nonl \tbf{Input:} Policy $\pith$, baseline $\vphi$, replay buffer $\DD$
\vspace*{0.05in}

Roll out trajectories $\bb = \cbrac{\t_1, \t_2, \ldots, \t_K}$ for $T$ time-steps each\\[0.05in]

Compute the returns $G(\t_k)$ and policy $\pith(\cdot | s_t;\ \t_k)$ $\forall\ t \leq T, k \leq K$\\[0.05in]

$\DD \la \DD \cup \bb$\\[0.025in]

On-policy update of $\pith$ using $\bb$; see~\cref{eq:p3o}\\[0.025in]

$\xi \la$ Poisson$(m)$\\[0.025in]

\For{$i \leq \xi$}
{   \vspace*{0.05in}
    $\bb_i \la$ sample mini-batch from $\DD$\\[0.025in]

    Estimate ESS and KL-divergence term using $\pith$ and stored policies $\log \mu(\cdot | s_t;\ \t_k)$ $\forall\ t \leq T, \t_k \in \bb_i$\\[0.025in]

    Off-policy and KL regularizer update of $\pith$ using $\bb_i$; see~\cref{eq:p3o}\\[0.025in]
}
\caption{One iteration of Policy-on Policy-off Policy Optimization (P3O)}
\label{alg:p3o}
\end{algorithm}
\DecMargin{0.2in}
}

\section{EXPERIMENTAL VALIDATION}
\label{s:expt}

This section demonstrates empirically that P3O with the ESS-based hyper-parameter choices from~\cref{s:approach} achieves, on-average, comparable performance to state-of-the-art algorithms. We evaluate the P3O algorithm against competitive baselines on the Atari-2600 benchmarks and MuJoCo continuous-control benchmarks.

\begin{figure*}[!ht]
\centering
\captionsetup[subfigure]{justification=centering}

\begin{subfigure}[t]{0.3\textwidth}
\centering
\includegraphics[width=\textwidth]{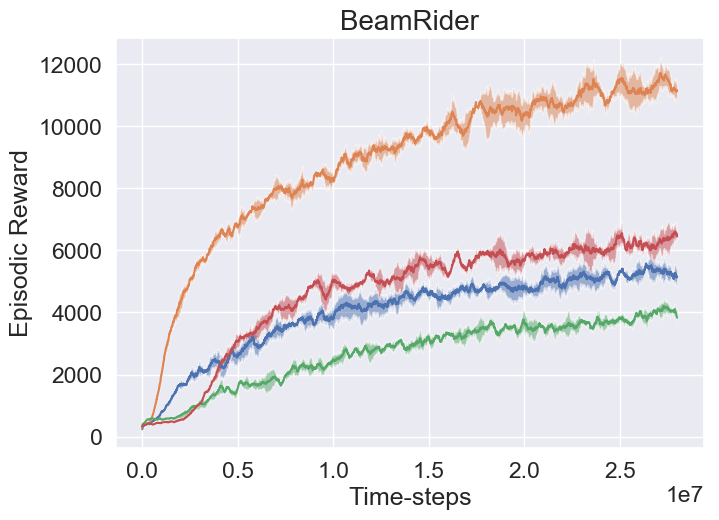}
\label{fig:atari:BeamRider}
\end{subfigure}
\hspace{0.1in}
\begin{subfigure}[t]{0.3\textwidth}
\centering
\includegraphics[width=\textwidth]{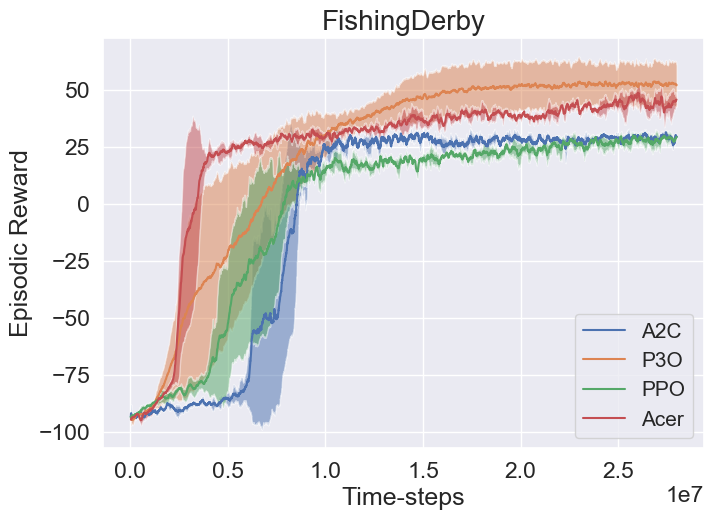}
\label{fig:atari:FishingDerby}
\end{subfigure}
\hspace{0.1in}
\begin{subfigure}[t]{0.3\textwidth}
\centering
\includegraphics[width=\textwidth]{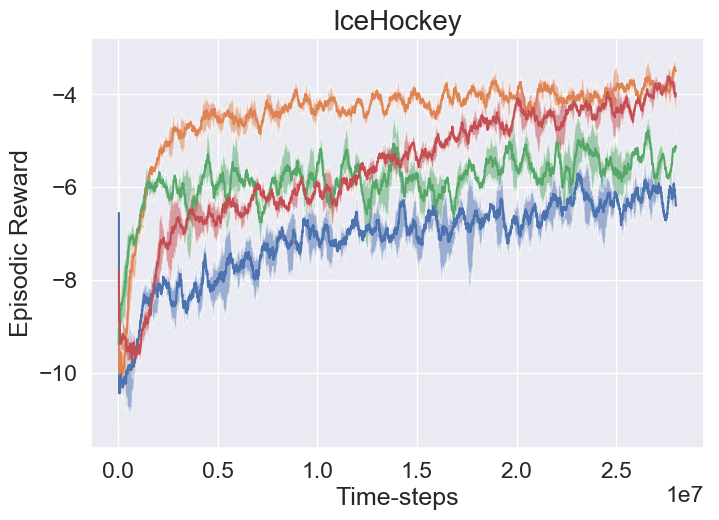}
\label{fig:atari:IceHockey}
\end{subfigure}

\begin{subfigure}[t]{0.3\textwidth}
\centering
\includegraphics[width=\textwidth]{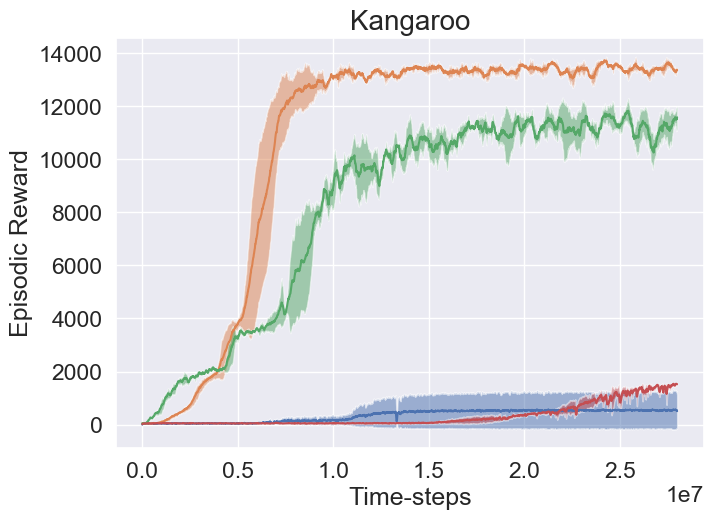}
\label{fig:atari:Kangaroo}
\end{subfigure}
\hspace{0.1in}
\begin{subfigure}[t]{0.3\textwidth}
\centering
\includegraphics[width=\textwidth]{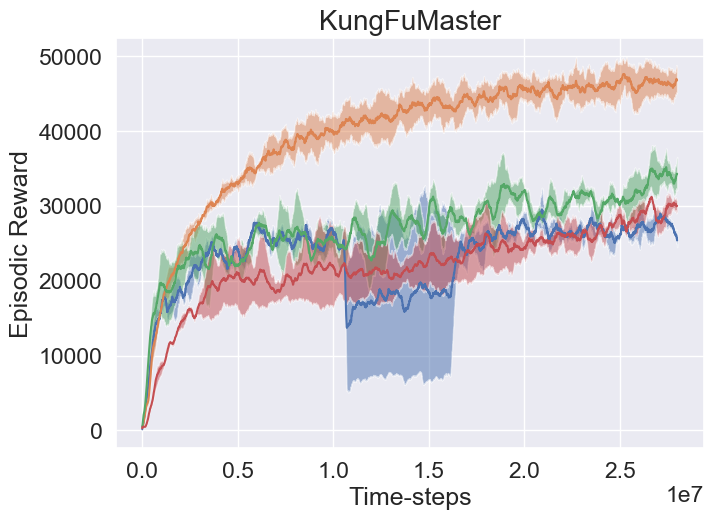}
\label{fig:atari:KungFuMaster}
\end{subfigure}
\hspace{0.1in}
\begin{subfigure}[t]{0.3\textwidth}
\centering
\includegraphics[width=\textwidth]{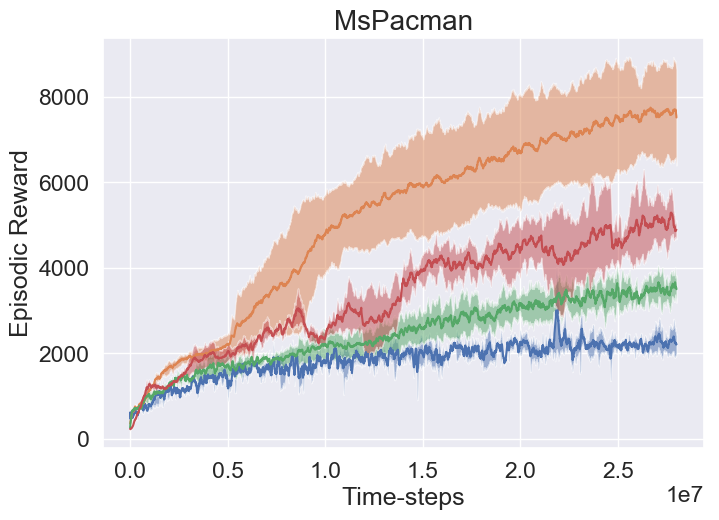}
\label{fig:atari:Ms. Pac-Man}
\end{subfigure}

\begin{subfigure}[t]{0.3\textwidth}
\centering
\includegraphics[width=\textwidth]{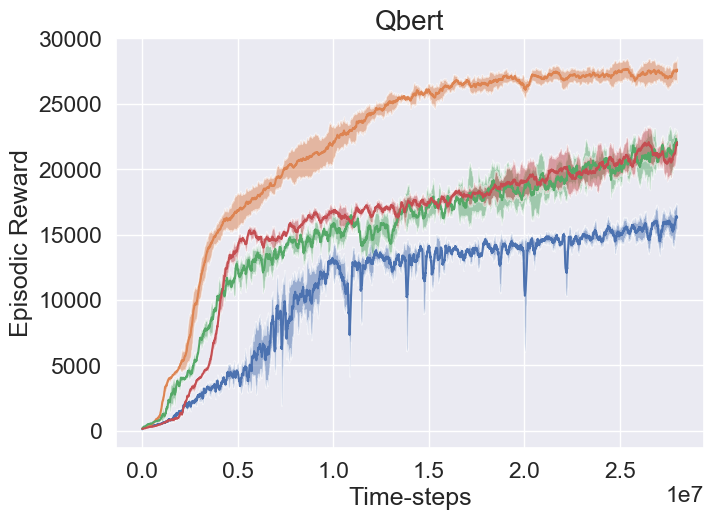}
\label{fig:atari:Qbert}
\end{subfigure}
\hspace{0.1in}
\begin{subfigure}[t]{0.3\textwidth}
\centering
\includegraphics[width=\textwidth]{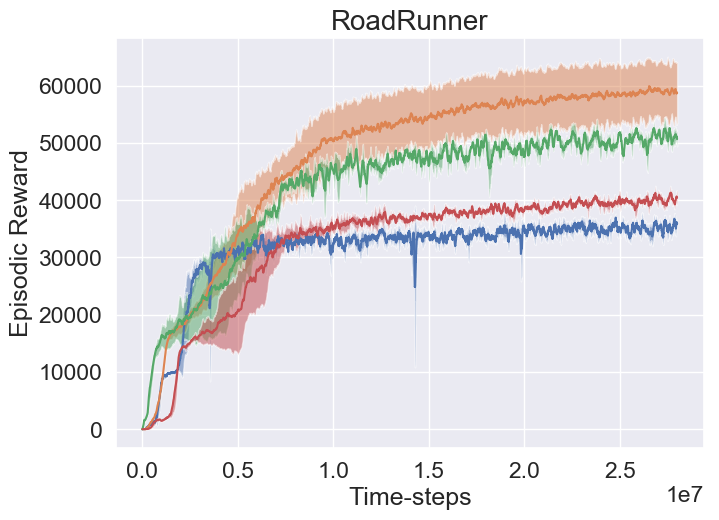}
\label{fig:atari:RoadRunner}
\end{subfigure}
\hspace{0.1in}
\begin{subfigure}[t]{0.3\textwidth}
\centering
\includegraphics[width=\textwidth]{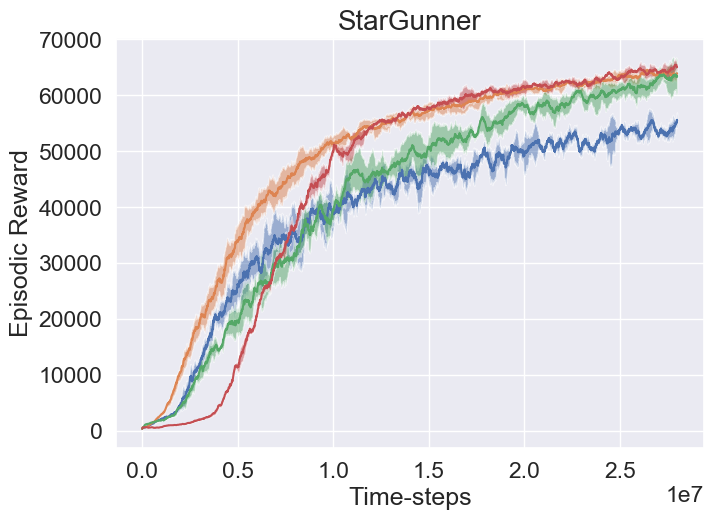}
\label{fig:atari:StarGunner}
\end{subfigure}

\caption{\tbf{Training curves for A2C (blue), ACER (red), PPO (green) and P3O (orange)} on some Atari games. See the Supplementary Material for similar plots on all Atari games.}
\label{fig:atari}
\end{figure*}

\subsection{Setup}
\label{ss:setup}

We compare P3O against three competitive baselines: the synchronous actor-critic architecture (A2C)~\citet{mnih2016asynchronous}, proximal policy optimization (PPO)~\citet{schulman2017proximal} and actor-critic with experience replay (ACER)~\citet{wang2016sample}. The first, A2C, is a standard baseline while PPO is a completely on-policy algorithm that is robust and has demonstrated good empirical performance. ACER combines on-policy updates with off-policy updates and is closest to P3O. We use the same network as that of~\citet{mnih2015human} for the Atari-2600 benchmark and a two-layer fully-connected network for MuJoCo tasks. The hyper-parameters are the same as those of the original authors of the above papers in order to be consistent and comparable to existing literature. We use implementations from OpenAI Baselines\footnote{\href{https://github.com/openai/baselines}{https://github.com/openai/baselines}}. We follow the evaluation protocol proposed by~\citep{MachadoAracde2017} and report the training returns for all experiments. More details are provided in the Supplementary Material.

\subsection{Results}

\tbf{Atari-2600 benchmark.} \cref{tab:atari} shows a comparison of P3O against the three baselines averaged over all the games in the Atari-2600 benchmark suite. We measure performance in two ways: (i) in terms of the final reward for each algorithm averaged over the last 100 episodes after 28M time-steps (112M frames of the game), and (ii) in terms of the reward at 40\% training time and 80\% training time averaged over 100 episodes. The latter compares different algorithms in terms of their sample efficiency. These results suggest that P3O is an efficient algorithm that improves upon competitive baselines both in terms of the final reward at the end of training and the reward obtained after a fixed number of samples. \cref{fig:atari} shows the reward curves for some of games; rewards and training curves for all games are provided in the Supplementary Material.

{
\renewcommand{\arraystretch}{1.2}
\begin{table}[!htpb]
\begin{center}
\small
\caption{\tbf{Number of Atari games ``won'' by each algorithm} measured by the average return over 100 episodes across three random seeds.}
{
\begin{tabular}{p{1.5cm} r r r}
\toprule
    \rowcolor{gray!15} Algorithm & Won & Won @ 40\% & Won @ 80\%\\
    \rowcolor{gray!15}  & & training time & training time\\
\toprule
    A2C & 0 & 0 & 0\\
    ACER & 13& 9 & 11 \\
    PPO & 9 & 8 & 10\\
    P3O & \tbf{27} & \tbf{32} & \tbf{28} \\
\bottomrule
\end{tabular}
}
\label{tab:atari}
\end{center}
\end{table}
}

Completely off-policy algorithms are a strong benchmark on Atari games. We therefore compare P3O with a few state of the art off-policy algorithms using published results by the original authors. P3O wins 32 games vs. 17 games won by DDQN~\citep{van2016deep}. P3O wins 18 games vs. 30 games won by C51~\citep{bellemare2017distributional}. P3O wins 26 games vs. 22 games won by SIL~\citep{oh2018self}. These off-policy algorithms use 200M frames and P3O's performance with 112M frames is comparable to them.

\tbf{MuJoCo continuous-control tasks.}
In addition to A2C and PPO, we also show a comparison to Q-Prop~\citep{gu2016q} and Interpolated Policy Gradients (IPG)~\citep{gu2017interpolated}; the returns for the latter are taken from the training curves in the original papers; they use 10M time-steps and 3 random seeds. The code of the original authors of ACER for MuJoCo is unavailable and we, as also others, were unsuccessful in getting ACER to train for continuous-control tasks. \cref{tab:mujoco} shows that P3O achieves better performance than strong baselines for continuous-control tasks such as A2C and PPO. It is also better than on-average than algorithms such as Q-Prop and IPG designed to combine off-policy and on-policy data. Note that Q-Prop/IPG were tuned by the original authors specifically for each task. In contrast, all hyper-parameters for P3O are fixed across the MuJoCo benchmarks. Training curves and results for more environments are in the Supplementary Material.

{
\renewcommand{\arraystretch}{1.35}
\begin{table}[!htpb]
\begin{center}
\small
\caption{\tbf{Average return on MuJoCo continuous-control tasks} after 3M time-steps of training on 10 seeds.}
\label{tab:mujoco}
{
\begin{tabular}{p{1.75cm} r r r r r r}
\toprule \rowcolor{gray!15}
Task         & A2C   & PPO       &  Q-Prop   & IPG   & P3O \\
\toprule
Half-Cheetah & 1907   & 2022     & 4178    & 4216   & \tbf{5052}\\
Walker      & 2015   & 2728     & 2832    & 1896    & \tbf{3771}\\
Hopper      & 1708   & 2245     & \tbf{2957}        & -    & 2334\\
Ant         & 1811   & 1616     & 3374    & 3943    & \tbf{4727}\\
Humanoid    & 720    & 530      & 1423    & 1651    & \tbf{2057}\\
\bottomrule
\end{tabular}
}
\end{center}
\end{table}
}

\section{RELATED WORK}
\label{s:related_work}

This work builds upon recent techniques that combine off-policy and on-policy updates in reinforcement learning. The closest to our approach is the ACER algorithm~\citep{wang2016sample}. It builds upon the off-policy actor-critic method\citep{degris2012off} and uses the Retrace operator~\citep{munos2016safe} to estimate an off-policy action-value function and constrains the candidate policy to be close to the running average of past policies using a linearized $\kl$-divergence penalty. P3O uses a biased variant of the ACER gradient and incorporates an explicit $\kl$ penalty in the objective.

The PGQL algorithm~\citep{o2016combining} uses an estimate of the action-value function of the target policy to combine on-policy updates with those obtained by minimizing the Bellman error. QProp~\citep{gu2016q} learns the action-value function using off-policy data which is used as a control variate for on-policy updates. The authors in~\citet{gu2017interpolated} propose the interpolated policy gradient (IPG) which takes a unified view of these algorithms. It directly combines on-policy and off-policy updates using a hyper-parameter and shows that, although such updates may be biased, the bias is bounded.

The key characteristic of the above algorithms is that they use hyper-parameters as a way to combine off-policy data with on-policy data. This is fragile in practice because different environments require different hyper-parameters. Moreover, the ideal hyper-parameters for combining data may change as training progresses; see~\cref{fig:effect_of_lambda}. For instance, the authors in~\citet{oh2018self} report poorer empirical results with ACER and prioritized replay as compared to vanilla actor-critic methods (A2C). The effective sample size heuristic (ESS) in P3O is a completely automatic, parameter-free way of combining off-policy data with on-policy data.

Policy gradient algorithms with off-policy data are not new. The importance sampling ratio has been commonly used by a number of authors such as~\citet{cao2005basic,levine2013guided}. Effective sample size is popularly used to measure the quality of importance sampling and to restrict the search space for parameter updates~\citep{jie2010connection,peshkin2002learning}. We exploit ESS to a similar end, it is an effective way to both control the contribution of the off-policy data and the deviation of the target policy from the behavior policy. Let us note there are a number of works that learn action-value functions using off-policy data, e.g.,\citet{wang2013backward,hausknecht2016policy,lehnert2015policy} that achieve varying degrees of success on reinforcement learning benchmarks.

Covariate shift and effective sample size have been studied extensively in the machine learning literature; see~\citet{robert2013monte,quionero2009dataset} for an elaborate treatment. These ideas have also been employed in reinforcement learning~\citep{kang2007demystifying,bang2005doubly,dudik2011doubly}. To the best of our knowledge, this paper is the first to use ESS for combining on-policy updates with off-policy updates.

\section{DISCUSSION}
\label{s:conc}

Sample complexity is the key inhibitor to translating the empirical performance of reinforcement learning algorithms from simulation to the real-world. Exploiting past, off-policy data to offset the high sample complexity of on-policy methods may be the key to doing so. Current approaches to combine the two using hyper-parameters are fragile. P3O is a simple, effective algorithm that uses the effective sample size (ESS) to automatically govern this combination. It demonstrates strong empirical performance across a variety of benchmarks. More generally, the discrepancy between the distribution of past data used to fit control variates and the data being gathered by the new policy lies at the heart of modern RL algorithms. The analysis of RL algorithms has not delved into this phenomenon. We believe this to be a promising avenue for future research.

\section{ACKNOWLEDGEMENTS}

The authors would like to acknowledge the support of Hang Zhang and Tong He from Amazon Web Services for the open-source implementation of P3O.

{
\footnotesize
\setlength{\bibsep}{0.5em}
\bibliographystyle{apalike}
\bibliography{refs}
}
\section*{Appendix}
\label{sec:appex}
\begin{appendix}

\section{Hyper-parameters for all experiments}

\begin{table}[!htp]
\small
\centering
\caption{\tbf{A2C hyper-parameters on Atari benchmark}}
\label{tab:hyper-a2c}
\begin{tabular}{p{4.5cm} r}
\toprule
\rowcolor{gray!15} Hyper-parameters & Value \\
\midrule
Architecture & conv ($32$-$8\times 8$-$4$)\\
& conv ($64$-$4\times 4$-$2$)\\
& conv ($64$-$3\times 1$-$1$)\\
& FC ($512$) \\
Learning rate & $7 \times 10^{-4}$ \\
Number of environments & 16 \\
Number of steps per iteration & 5 \\
Entropy regularization ($\a$) & 0.01 \\
Discount factor ($\g$) & $0.99$ \\
Value loss Coefficient & $0.5$ \\
Gradient norm clipping coefficient & $0.5$ \\
Random Seeds & $\{0\dots2\}$ \\
\bottomrule
\end{tabular}
\end{table}

\begin{table}[!thpb]
\small
\centering
\caption{\tbf{ACER hyper-parameters on Atari benchmark}}
\label{tab:hyper-acer}
\begin{tabular}{p{4.5cm} r}
\toprule
\rowcolor{gray!15} Hyper-parameters & Value \\
\midrule
Architecture & Same as A2C \\
Replay Buffer size & $5 \times 10^4$ \\
Learning rate & $7 \times 10^{-4}$ \\
Number of environments & 16 \\
Number of steps per iteration & 20 \\
Entropy regularization ($\a$) & 0.01 \\
Number of training epochs per update & 4 \\
Discount factor ($\g$) & $0.99$ \\
Value loss Coefficient & $0.5$ \\
importance weight clipping factor & $10$ \\
Gradient norm clipping coefficient & $0.5$ \\
Momentum factor in the Polyak & $0.99$ \\
Max. KL between old \& updated policy & $1$ \\
Use Trust region & True \\
Random Seeds & $\{0\dots2\}$ \\
\bottomrule
\end{tabular}
\end{table}

\begin{table}[!thpb]
\small
\centering
\caption{\tbf{PPO hyper-parameters on Atari benchmark}}
\label{tab:hyper-ppo}
\begin{tabular}{p{4.5cm} r}
\toprule
\rowcolor{gray!15} Hyper-parameters & Value \\
\midrule
Architecture & Same as A2C \\
Learning rate & $7 \times 10^{-4}$ \\
Number of environments & 8 \\
Number of steps per iteration & 128 \\
Entropy regularization ($\a$) & 0.01 \\
Number of training epochs per update & 4 \\
Discount factor ($\g$) & $0.99$ \\
Value loss Coefficient & $0.5$ \\
Gradient norm clipping coefficient & $0.5$ \\
Advantage estimation discounting factor ($\t$) & $0.95$ \\
Random Seeds & $\{0\dots2\}$ \\
\bottomrule
\end{tabular}
\end{table}

\renewcommand{\arraystretch}{1.1}
\begin{table}[!thpb]
\small
\centering
\caption{\tbf{P3O hyper-parameters on Atari benchmark}}
\label{tab:hyper-p3o}
\begin{tabular}{p{4.5cm} r}
\toprule
\rowcolor{gray!15} Hyper-parameters & Value \\
\midrule
Architecture & Same as A2C \\
Learning rate & $7 \times 10^{-4}$ \\
Replay Buffer size & $5 \times 10^4$ \\
Number of environments & 16 \\
Number of steps per iteration & 16 \\
Entropy regularization ($\a$) & 0.01 \\
Off policy updates per iteration ($\xi$) & Poisson(2) \\
Burn-in period & $15 \times 10^3$ \\
Samples from replay buffer  & $6$ \\
Discount factor ($\g$) & $0.99$ \\
Value loss Coefficient & $0.5$ \\
Gradient norm clipping coefficient & $0.5$ \\
Advantage estimation discounting factor ($\tau$) & $0.95$ \\
Random Seeds & $\{0\dots2\}$ \\
\bottomrule
\end{tabular}
\end{table}

\begin{table}[!htpb]
\small
\centering
\caption{\tbf{P3O hyper-parameters for MuJoCo tasks}}
\label{tab:hyper-p3o}

\begin{tabular}{p{4.5cm} r}
\toprule
\rowcolor{gray!15} Hyper-parameters & Value \\
\midrule
Architecture & FC(100) - FC(100)\\
Learning rate & $3 \times 10^{-4}$ \\
Replay Buffer size & $5 \times 10^3$ \\
Number of environments & 2 \\
Number of steps per iteration & 64 \\
Entropy regularization ($\a$) & 0.0 \\
Off policy updates per iteration ($\xi$) & Poisson(3) \\
Burn-in period & 2500 \\
Number of samples from replay buffer  & $15$ \\
Discount factor ($\g$) & $0.99$ \\
Value loss Coefficient & $0.5$ \\
Gradient norm clipping coefficient & $0.5$ \\
Advantage estimation discounting factor ($\tau$) & $0.95$ \\
Random Seeds & $\{0\dots9\}$ \\
\bottomrule
\end{tabular}
\end{table}

\begin{table}[!htpb]
\small
\centering
\caption{\tbf{A2C (and A2C with GAE) hyper-parameters on MuJoCo tasks}}
\label{tab:hyper-a2c}
\begin{tabular}{p{4.5cm} r}
\toprule
\rowcolor{gray!15} Hyper-parameters & Value \\
\midrule
Architecture & FC(64) - FC(64) \\
Learning rate & $ 13 \times 10^{-3}$ \\
Number of environments & 8 \\
Number of steps per iteration & 32 \\
Entropy regularization ($\a$) & 0.0 \\
Discount factor ($\g$) & $0.99$ \\
Value loss Coefficient & $0.5$ \\
Gradient norm clipping coefficient & $0.5$ \\
Random Seeds & $\{0\dots9\}$ \\
\bottomrule
\end{tabular}
\end{table}

\begin{table}[!htpb]
\small
\centering
\caption{\tbf{PPO hyper-parameters on MuJoCo tasks}}
\label{tab:hyper-ppo-m}
\begin{tabular}{p{4.5cm} r}
\toprule
\rowcolor{gray!15} Hyper-parameters & Value \\
\midrule
Architecture & FC(64) - FC(64) \\
Learning rate & $3 \times 10^{-4}$ \\
Number of environments & 1 \\
Number of steps per iteration & 2048 \\
Entropy regularization ($\a$) & 0.0 \\
Number of training epochs per update & 10 \\
Discount factor ($\g$) & $0.99$ \\
Value loss Coefficient & $0.5$ \\
Gradient norm clipping coefficient & $0.5$ \\
Advantage estimation discounting factor ($\t$) & $0.95$ \\
Random Seeds & $\{0\dots9\}$ \\
\bottomrule
\end{tabular}
\end{table}

\section{Comparisons with baseline algorithms}
\label{supp:comparisons}

\begin{figure*}[t]
\centering
\captionsetup[subfigure]{justification=centering}
\begin{subfigure}[t]{0.315\textwidth}
\centering
\includegraphics[width=\textwidth]{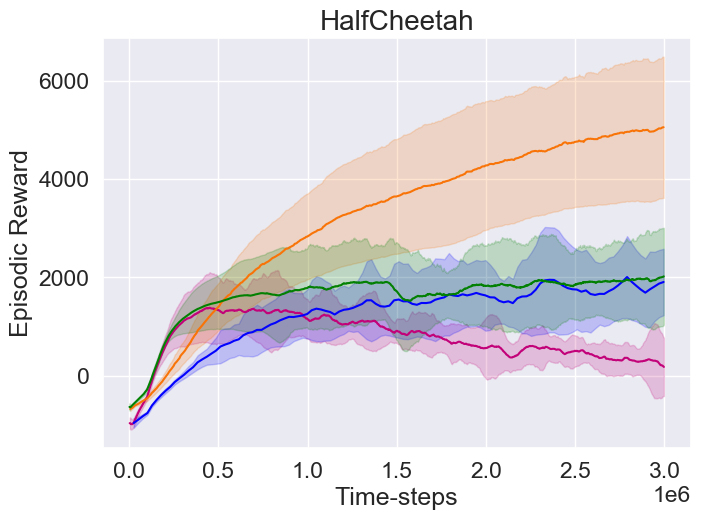}
\label{fig:HalfCheetah}
\end{subfigure}%
~
\begin{subfigure}[t]{0.32\textwidth}
\centering
\includegraphics[width=\textwidth]{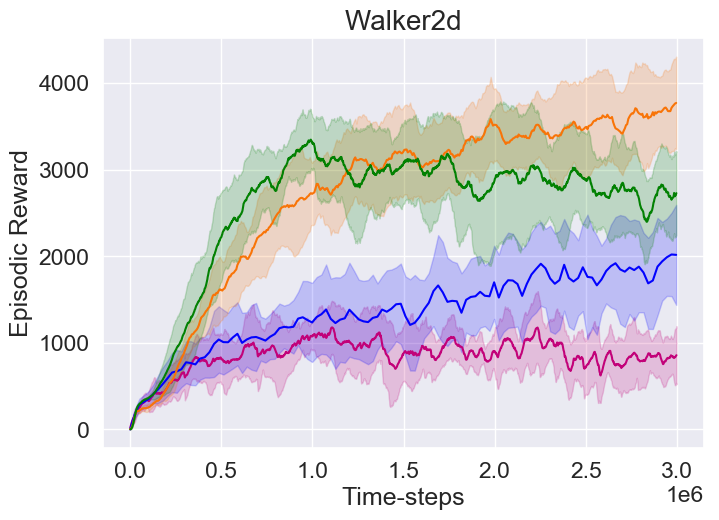}
\label{fig:Walker2d}
\end{subfigure}%
~
\begin{subfigure}[t]{0.32\textwidth}
\centering
\includegraphics[width=\textwidth]{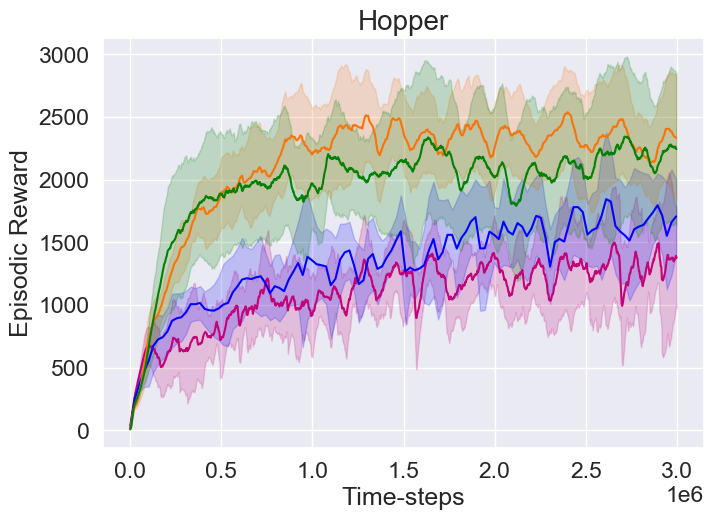}
\label{fig:Hopper}
\end{subfigure}%

\begin{subfigure}[t]{0.32\textwidth}
\centering
\includegraphics[width=\textwidth]{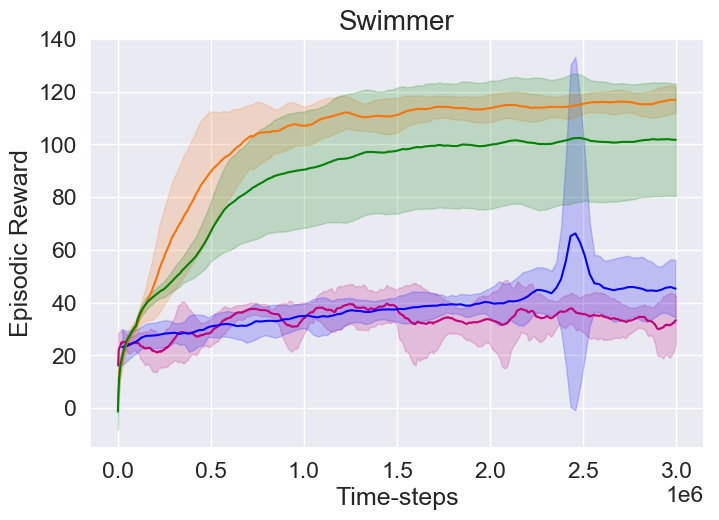}
\label{fig:Swimmer}
\end{subfigure}%
~
\begin{subfigure}[t]{0.32\textwidth}
\centering
\includegraphics[width=\textwidth]{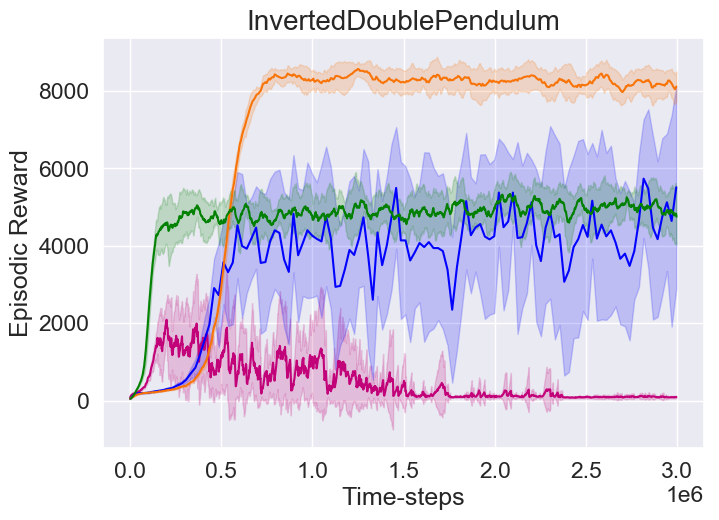}
\label{fig:InvertedDoublePendulum}
\end{subfigure}%
~
\begin{subfigure}[t]{0.32\textwidth}
\centering
\includegraphics[width=\textwidth]{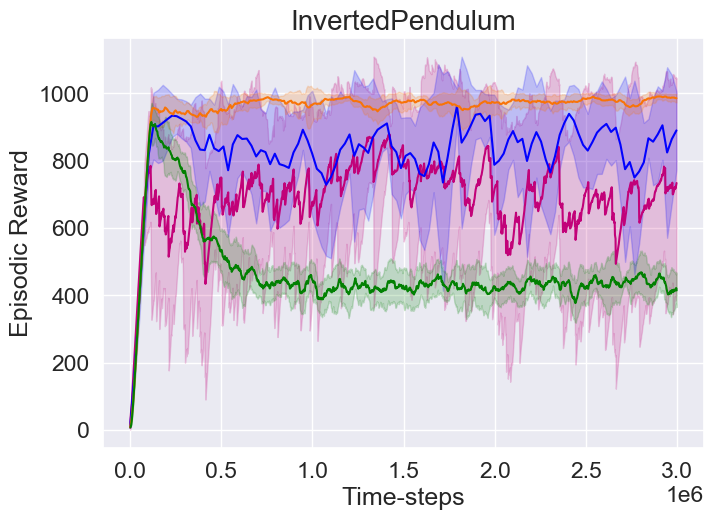}
\label{fig:InvertedPendulum}
\end{subfigure}%

\begin{subfigure}[t]{0.32\textwidth}
\centering
\includegraphics[width=\textwidth]{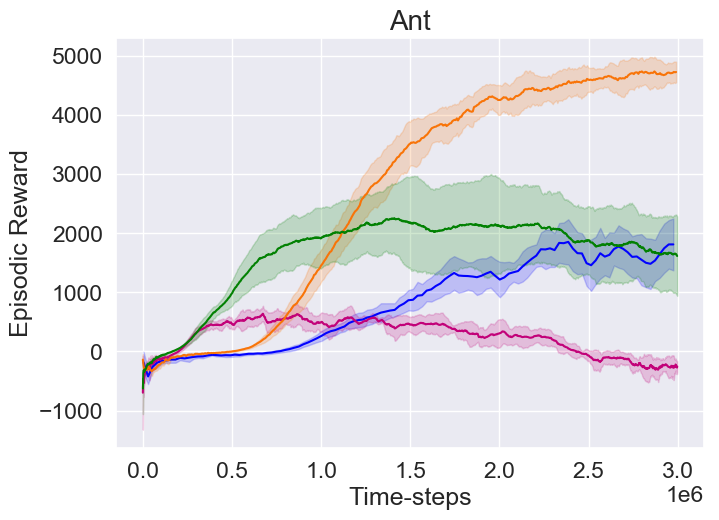}
\label{fig:Ant}
\end{subfigure}%
~
\begin{subfigure}[t]{0.32\textwidth}
\centering
\includegraphics[width=\textwidth]{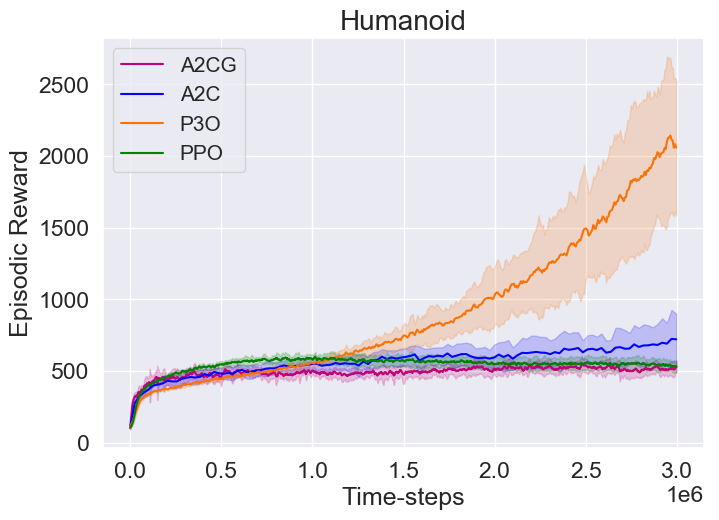}
\label{fig:Humanoid}
\end{subfigure}%

\caption{\tbf{Training curves of A2C (blue), A2CG [A2C with GAE] (magenta), PPO (green) and P3O (orange) on 8 MuJoCo environments.}}
\label{fig:mujoco}
\end{figure*}

{
\renewcommand{\arraystretch}{1.35}
\begin{table}[!htpb]
\begin{center}
\small
\caption{\tbf{Returns on MuJoCo continuous-control tasks after 3M time-steps of training and 10 random seeds}.}
\label{tab:supp:mujoco}
{
\begin{tabular}{p{1.6cm}rrrr}
\toprule
    \rowcolor{gray!15} Games & A2CG & A2C  & PPO &  P3O \\
\toprule
Half-Cheetah  & 181.46  &  1907.42 & 2022.14 & \tbf{5051.58}\\
Walker  & 855.62  & 2015.15 & 2727.93 & \tbf{3770.86}\\
Hopper  & 1377.07  & 1708.22 & 2245.03 & \tbf{2334.32}\\
Swimmer  & 33.33  & 45.27 & 101.71 & \tbf{116.87}\\
Inverted Double Pendulum  & 90.09  & 5510.71 & 4750.69 & \tbf{8114.05}\\
Inverted Pendulum  & 733.34  & 889.61 & 414.49 & \tbf{985.14}\\
Ant  & -253.54  & 1811.29 & 1615.55 & \tbf{4727.34}\\
Humanoid  & 530.12  & 720.38 & 530.13 & \tbf{2057.17}\\
\bottomrule
\end{tabular}
}
\end{center}
\end{table}
}

\clearpage
\begin{table*}[t]
\begin{center}
\linespread{1.2}
\small
\caption{\tbf{Returns of agents on 49 Atari-2600 games after 28M timesteps (112M frames) of training.}}
\label{tab:atari}
{
\begin{tabular}{|p{4cm} |c  | c  | c  | c |}
\toprule
    \rowcolor{gray!15} Games & A2C  & ACER & PPO & P3O \\
\toprule
Alien  & $1425.00$ & $2436.20$ & $2260.43$ & $\mathbf{3124.80}$\\
Amidar  & $439.43$ & $1393.24$ & $1062.73$ & $\mathbf{1787.40}$\\
Assault  & $3897.73$ & $\mathbf{6996.46}$ & $5941.23$ & $6222.27$\\
Asterix  & $12272.50$ & $24414.00$ & $7574.33$ & $\mathbf{25997.00}$\\
Asteroids  & $2052.27$ & $1874.83$ & $2147.33$ & $\mathbf{2483.30}$\\
Atlantis  & $2847251.67$ & $2832752.33$ & $2647593.67$ & $\mathbf{3077883.00}$\\
BankHeist  & $910.43$ & $\mathbf{1281.60}$ & $1236.90$ & $864.03$\\
BattleZone  & $6250.00$ & $10726.67$ & $\mathbf{22856.67}$ & $12793.33$\\
BeamRider  & $5149.29$ & $6486.07$ & $3834.01$ & $\mathbf{11163.49}$\\
Bowling  & $24.19$ & $\mathbf{38.61}$ & $31.75$ & $27.04$\\
Boxing  & $0.21$ & $99.33$ & $98.06$ & $\mathbf{99.44}$\\
Breakout  & $403.25$ & $\mathbf{474.81}$ & $328.80$ & $351.81$\\
Centipede  & $3722.24$ & $6755.41$ & $4530.21$ & $\mathbf{8615.36}$\\
ChopperCommand  & $1389.67$ & $\mathbf{10376.00}$ & $9504.33$ & $8878.33$\\
CrazyClimber  & $111418.67$ & $136527.67$ & $118501.00$ & $\mathbf{168115.00}$\\
DemonAttack  & $65766.90$ & $181679.27$ & $37026.17$ & $\mathbf{331454.95}$\\
DoubleDunk  & $-17.86$ & $-8.37$ & $-6.29$ & $\mathbf{-3.83}$\\
Enduro  & $0.00$ & $0.00$ & $\mathbf{1092.52}$ & $0.00$\\
FishingDerby  & $29.54$ & $45.74$ & $29.34$ & $\mathbf{52.07}$\\
Freeway  & $0.00$ & $0.00$ & $\mathbf{32.83}$ & $0.00$\\
Frostbite  & $269.87$ & $304.23$ & $\mathbf{1266.73}$ & $312.13$\\
Gopher  & $3923.13$ & $\mathbf{99855.53}$ & $6451.07$ & $29603.60$\\
Gravitar  & $377.33$ & $387.00$ & $\mathbf{1042.67}$ & $987.50$\\
IceHockey  & $-6.39$ & $-3.97$ & $-5.11$ & $\mathbf{-3.50}$\\
Jamesbond  & $453.83$ & $457.50$ & $\mathbf{683.67}$ & $475.00$\\
Kangaroo  & $507.33$ & $1524.67$ & $11583.67$ & $\mathbf{13360.67}$\\
Krull  & $8935.40$ & $\mathbf{9115.73}$ & $8718.40$ & $7812.03$\\
KungFuMaster  & $25395.00$ & $30002.33$ & $34292.00$ & $\mathbf{46761.67}$\\
MontezumaRevenge  & $0.00$ & $0.00$ & $0.00$ & $\mathbf{805.33}$\\
MsPacman  & $2220.63$ & $4892.33$ & $3502.20$ & $\mathbf{7516.21}$\\
NameThisGame  & $5977.63$ & $\mathbf{15640.83}$ & $6011.03$ & $9232.70$\\
Pitfall  & $-65.50$ & $-7.64$ & $\mathbf{-1.94}$ & $-7.40$\\
Pong  & $20.21$ & $20.80$ & $20.69$ & $\mathbf{20.95}$\\
PrivateEye  & $49.24$ & $\mathbf{99.00}$ & $97.33$ & $92.61$\\
Qbert  & $16289.08$ & $22051.67$ & $21830.17$ & $\mathbf{27619.33}$\\
Riverraid  & $9680.33$ & $\mathbf{17794.03}$ & $11841.03$ & $13966.67$\\
RoadRunner  & $35918.33$ & $40428.67$ & $50663.33$ & $\mathbf{58728.00}$\\
Robotank  & $4.30$ & $4.89$ & $18.54$ & $\mathbf{33.69}$\\
Seaquest  & $1485.33$ & $1739.87$ & $\mathbf{1953.53}$ & $1851.87$\\
SpaceInvaders  & $1894.02$ & $\mathbf{3140.17}$ & $2124.57$ & $2699.33$\\
StarGunner  & $55469.33$ & $\mathbf{65005.00}$ & $63375.67$ & $63905.00$\\
Tennis  & $-22.22$ & $-11.26$ & $-6.72$ & $\mathbf{-5.27}$\\
TimePilot  & $3359.00$ & $7012.00$ & $7535.67$ & $\mathbf{10789.00}$\\
Tutankham  & $105.28$ & $\mathbf{291.09}$ & $206.42$ & $268.24$\\
UpNDown  & $30932.20$ & $159642.17$ & $173208.13$ & $\mathbf{279107.53}$\\
Venture  & $0.00$ & $0.00$ & $\mathbf{0.00}$ & $0.00$\\
VideoPinball  & $21061.76$ & $373803.36$ & $220680.47$ & $\mathbf{377935.99}$\\
WizardOfWor  & $1256.33$ & $2973.00$ & $5744.67$ & $\mathbf{10637.33}$\\
Zaxxon  & $17.00$ & $89.33$ & $8872.67$ & $\mathbf{16801.33}$\\
\bottomrule
\end{tabular}
}
\end{center}
\end{table*}

\begin{figure*}
\centering
\captionsetup[subfigure]{justification=centering}
\begin{subfigure}[t]{0.165\textwidth}
\centering
\includegraphics[width=\textwidth]{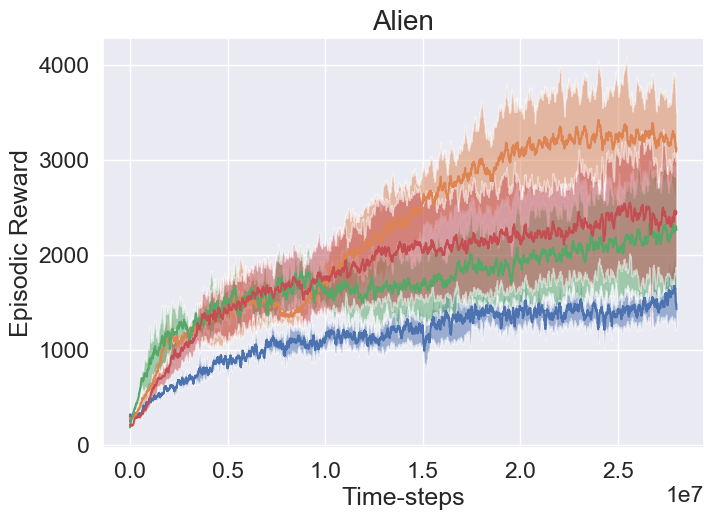}
\label{fig:Alien}
\end{subfigure}%
~
\begin{subfigure}[t]{0.165\textwidth}
\centering
\includegraphics[width=\textwidth]{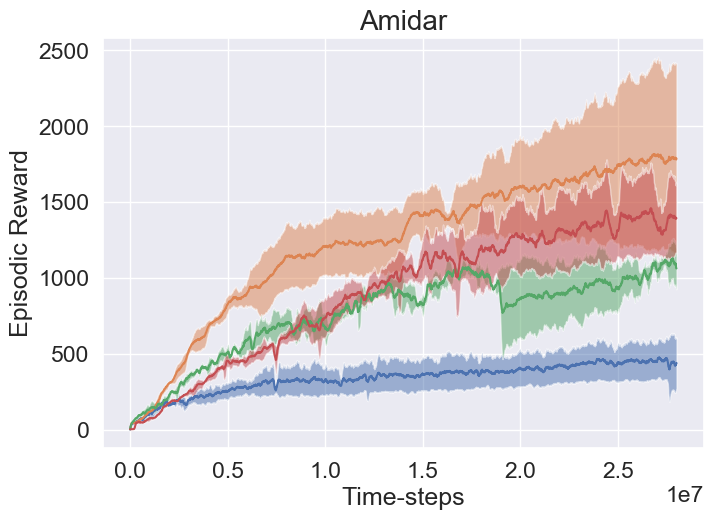}
\label{fig:Amidar}
\end{subfigure}%
~
\begin{subfigure}[t]{0.165\textwidth}
\centering
\includegraphics[width=\textwidth]{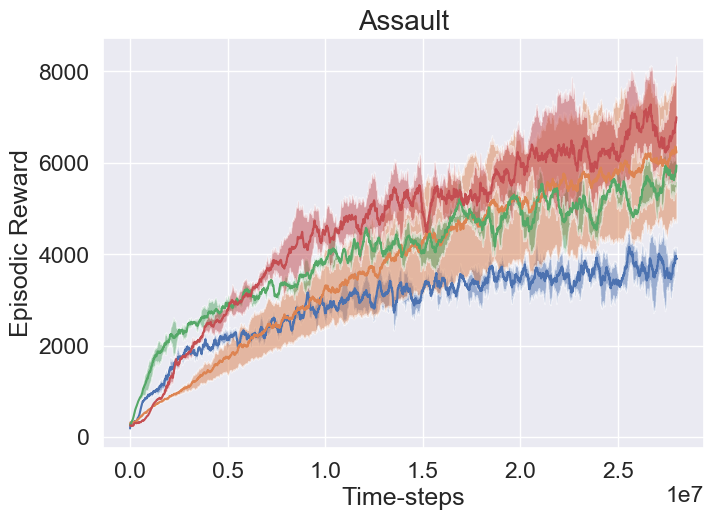}
\label{fig:Assault}
\end{subfigure}%
~
\begin{subfigure}[t]{0.165\textwidth}
\centering
\includegraphics[width=\textwidth]{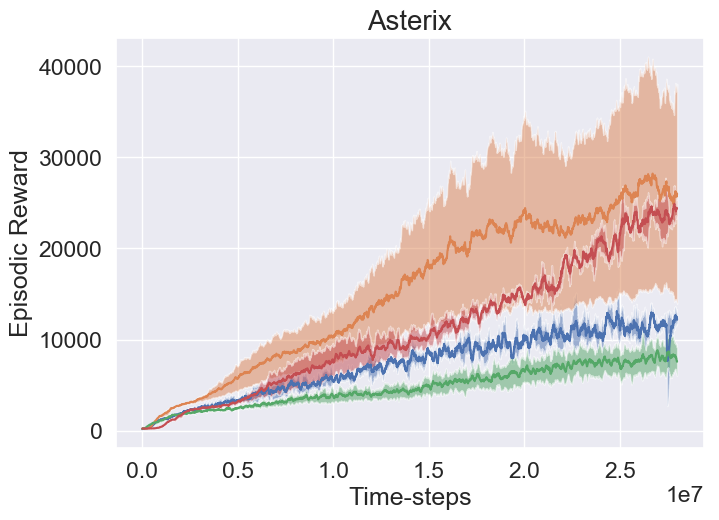}
\label{fig:Asterix}
\end{subfigure}%
~
\begin{subfigure}[t]{0.165\textwidth}
\centering
\includegraphics[width=\textwidth]{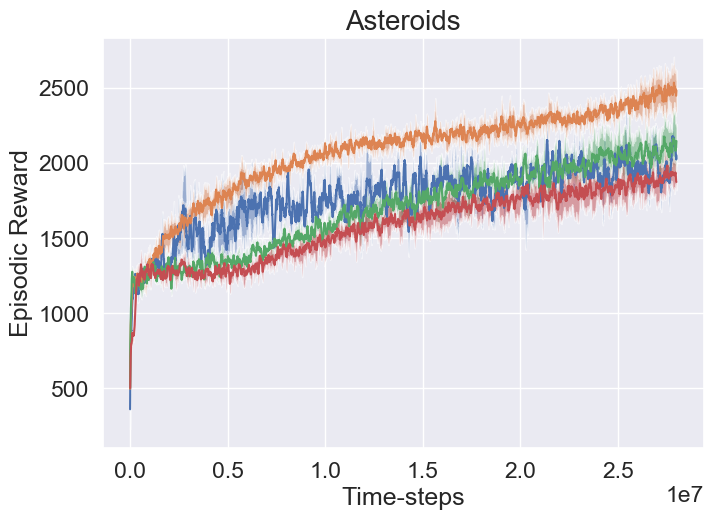}
\label{fig:Asteroids}
\end{subfigure}%

\begin{subfigure}[t]{0.165\textwidth}
\centering
\includegraphics[width=\textwidth]{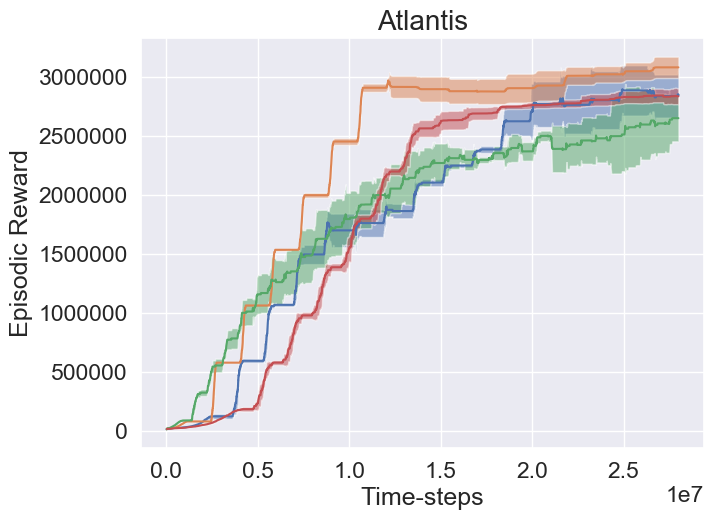}
\label{fig:Atlantis}
\end{subfigure}%
~
\begin{subfigure}[t]{0.165\textwidth}
\centering
\includegraphics[width=\textwidth]{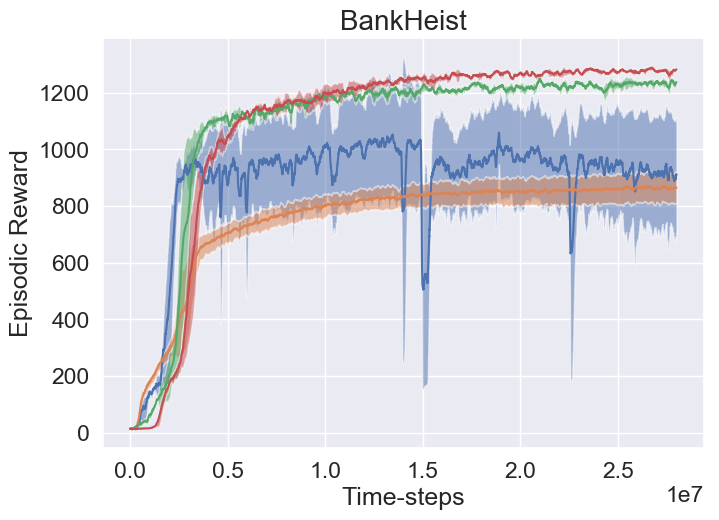}
\label{fig:BankHeist}
\end{subfigure}%
~
\begin{subfigure}[t]{0.165\textwidth}
\centering
\includegraphics[width=\textwidth]{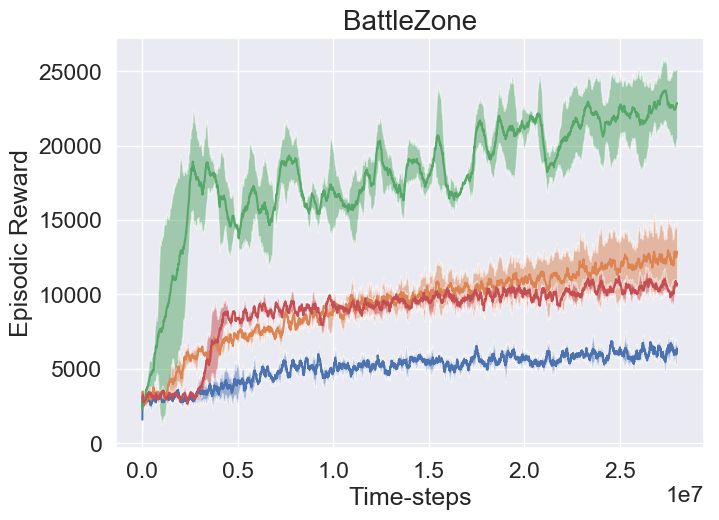}
\label{fig:BattleZone}
\end{subfigure}%
~
\begin{subfigure}[t]{0.165\textwidth}
\centering
\includegraphics[width=\textwidth]{BeamRider}
\label{fig:BeamRider}
\end{subfigure}%
~
\begin{subfigure}[t]{0.165\textwidth}
\centering
\includegraphics[width=\textwidth]{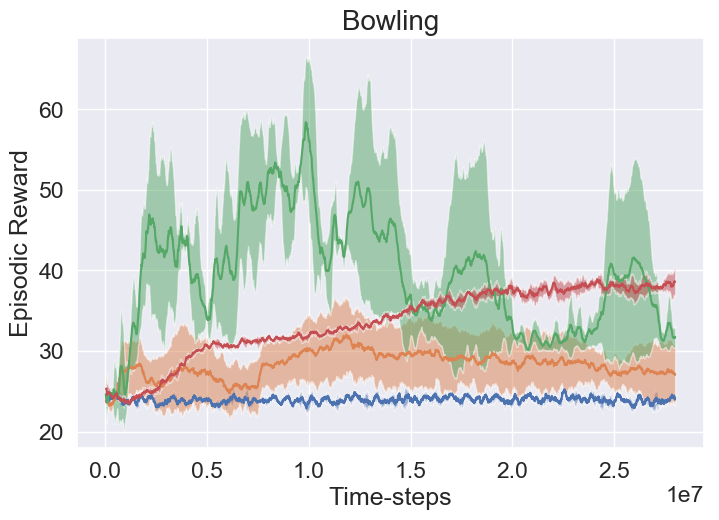}
\label{fig:Bowling}
\end{subfigure}%

\begin{subfigure}[t]{0.165\textwidth}
\centering
\includegraphics[width=\textwidth]{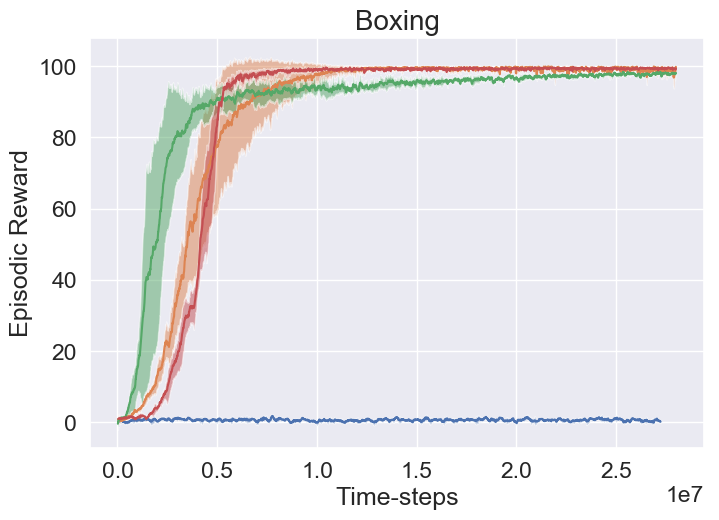}
\label{fig:Boxing}
\end{subfigure}%
~
\begin{subfigure}[t]{0.165\textwidth}
\centering
\includegraphics[width=\textwidth]{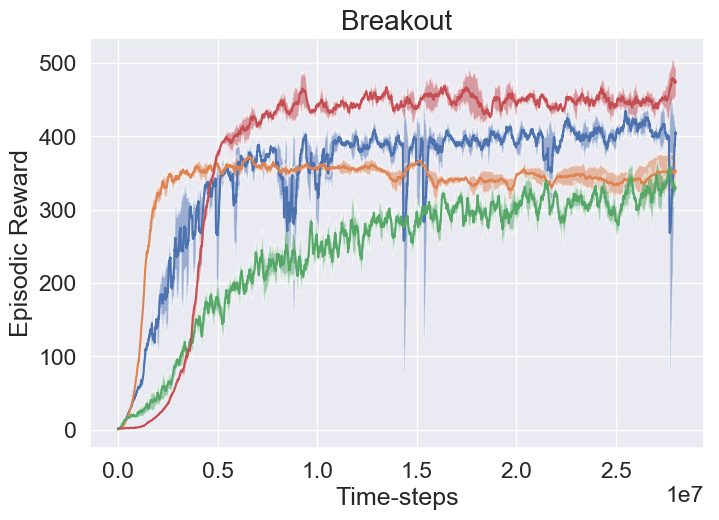}
\label{fig:Breakout}
\end{subfigure}%
~
\begin{subfigure}[t]{0.165\textwidth}
\centering
\includegraphics[width=\textwidth]{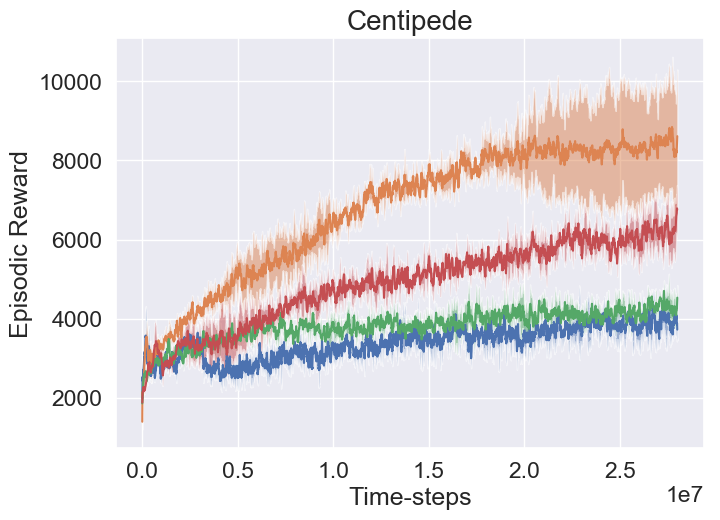}
\label{fig:Centipede}
\end{subfigure}%
~
\begin{subfigure}[t]{0.165\textwidth}
\centering
\includegraphics[width=\textwidth]{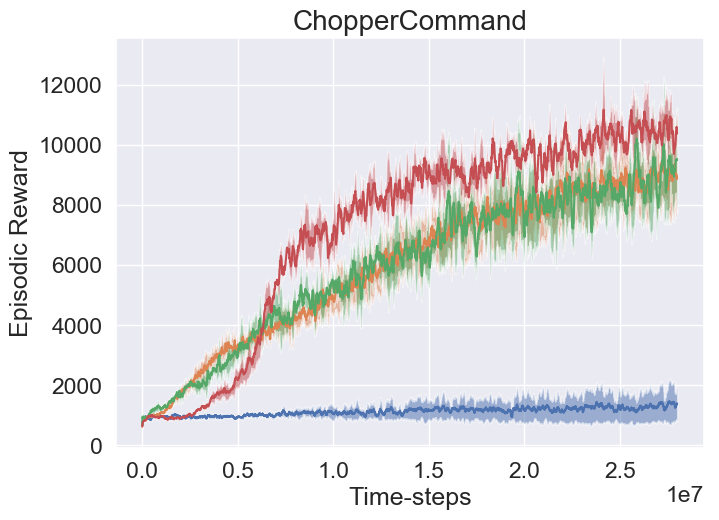}
\label{fig:ChopperCommand}
\end{subfigure}%
~
\begin{subfigure}[t]{0.165\textwidth}
\centering
\includegraphics[width=\textwidth]{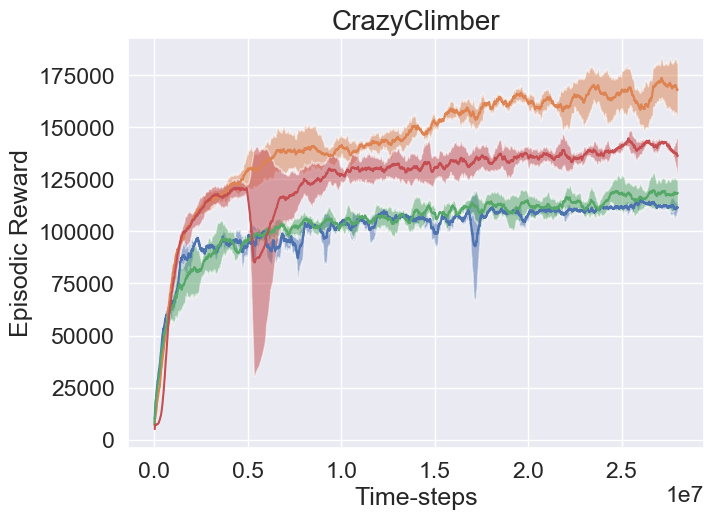}
\label{fig:CrazyClimber}
\end{subfigure}%

\begin{subfigure}[t]{0.165\textwidth}
\centering
\includegraphics[width=\textwidth]{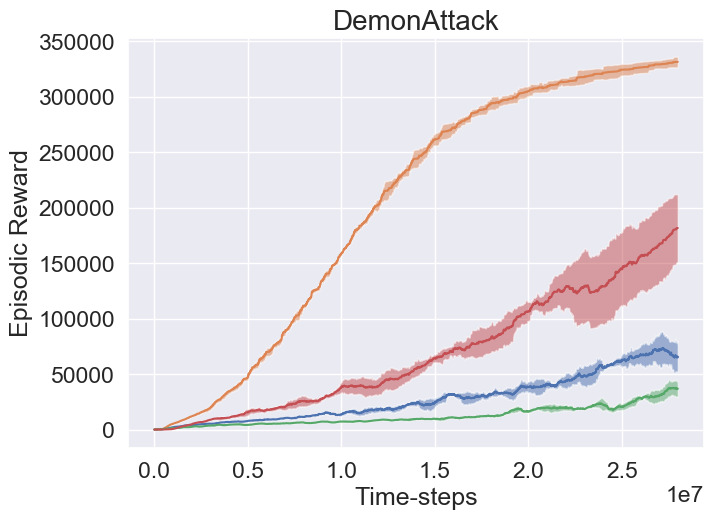}
\label{fig:DemonAttack}
\end{subfigure}%
~
\begin{subfigure}[t]{0.165\textwidth}
\centering
\includegraphics[width=\textwidth]{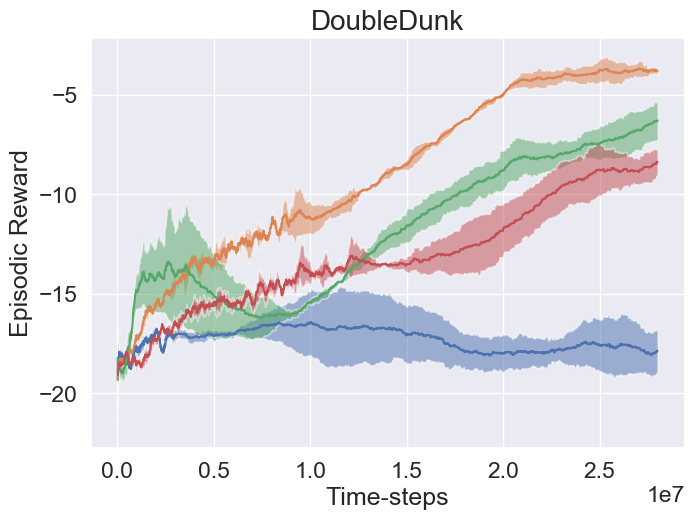}
\label{fig:DoubleDunk}
\end{subfigure}%
~
\begin{subfigure}[t]{0.165\textwidth}
\centering
\includegraphics[width=\textwidth]{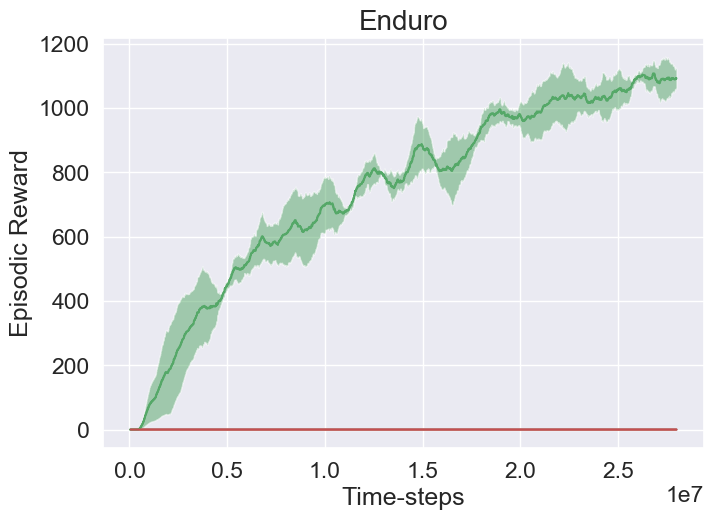}
\label{fig:Enduro}
\end{subfigure}%
~
\begin{subfigure}[t]{0.165\textwidth}
\centering
\includegraphics[width=\textwidth]{FishingDerby}
\label{fig:FishingDerby}
\end{subfigure}%
~
\begin{subfigure}[t]{0.165\textwidth}
\centering
\includegraphics[width=\textwidth]{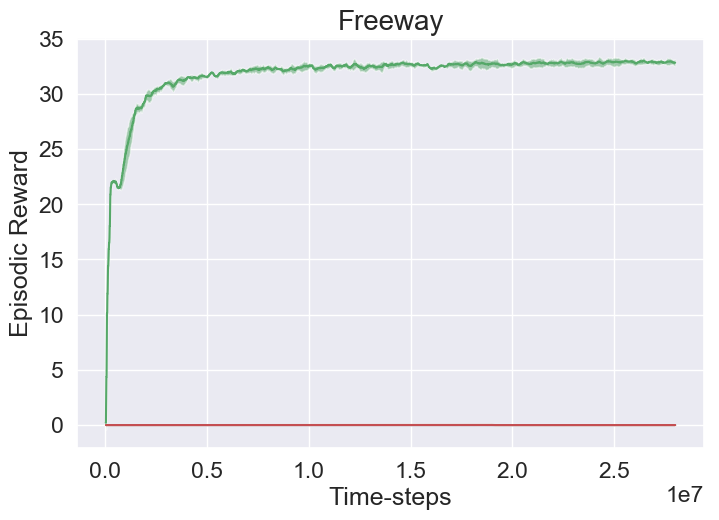}
\label{fig:Freeway}
\end{subfigure}%

\begin{subfigure}[t]{0.165\textwidth}
\centering
\includegraphics[width=\textwidth]{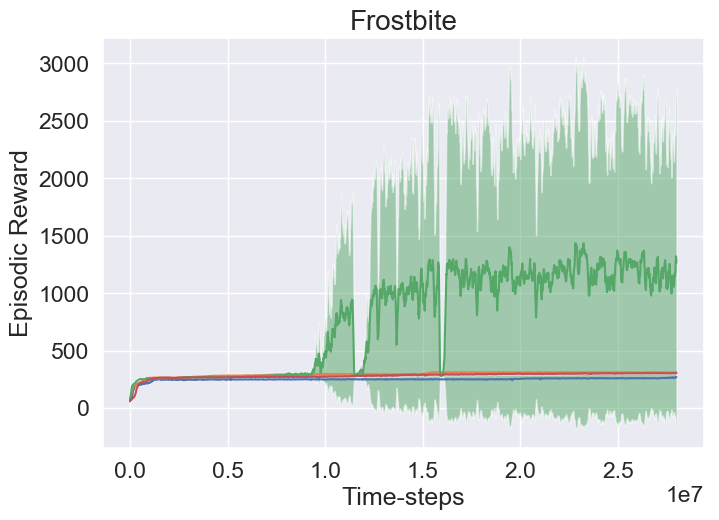}
\label{fig:Frostbite}
\end{subfigure}%
~
\begin{subfigure}[t]{0.165\textwidth}
\centering
\includegraphics[width=\textwidth]{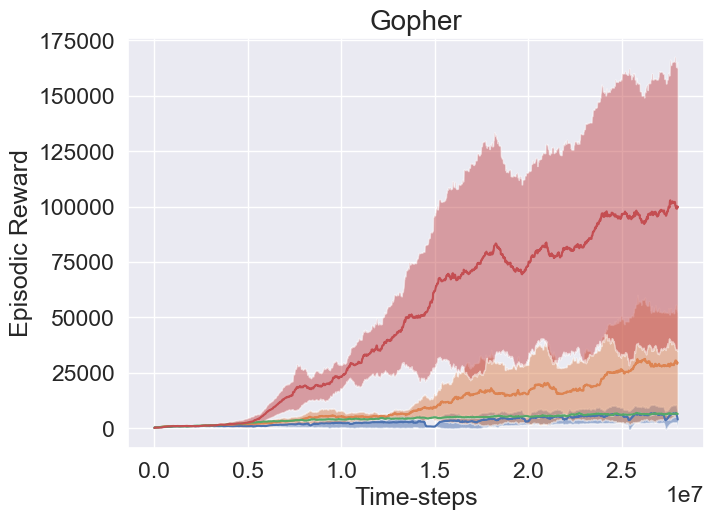}
\label{fig:Gopher}
\end{subfigure}%
~
\begin{subfigure}[t]{0.165\textwidth}
\centering
\includegraphics[width=\textwidth]{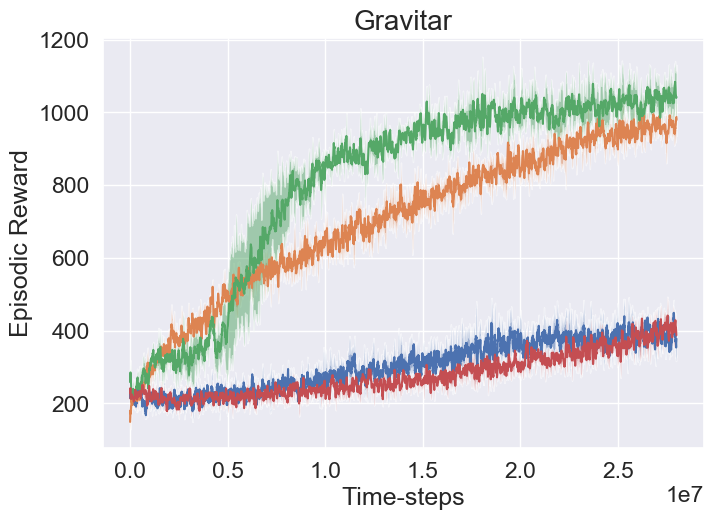}
\label{fig:Gravitar}
\end{subfigure}%
~
\begin{subfigure}[t]{0.165\textwidth}
\centering
\includegraphics[width=\textwidth]{IceHockey}
\label{fig:IceHockey}
\end{subfigure}%
~
\begin{subfigure}[t]{0.165\textwidth}
\centering
\includegraphics[width=\textwidth]{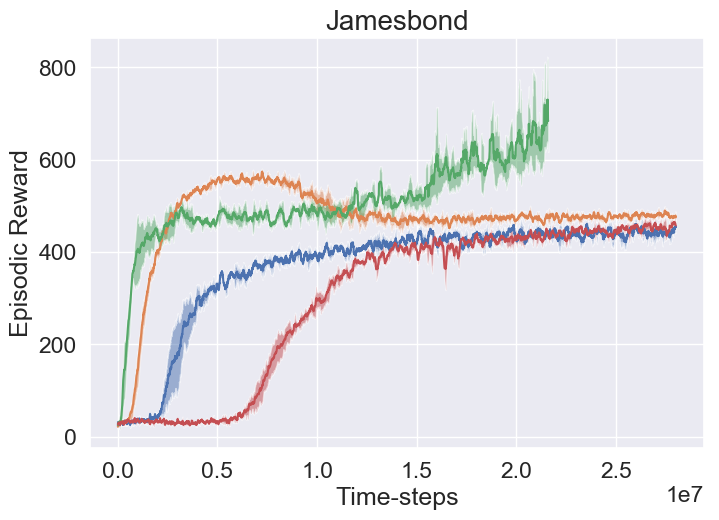}
\label{fig:Jamesbond}
\end{subfigure}%

\begin{subfigure}[t]{0.165\textwidth}
\centering
\includegraphics[width=\textwidth]{Kangaroo}
\label{fig:Kangaroo}
\end{subfigure}%
~
\begin{subfigure}[t]{0.165\textwidth}
\centering
\includegraphics[width=\textwidth]{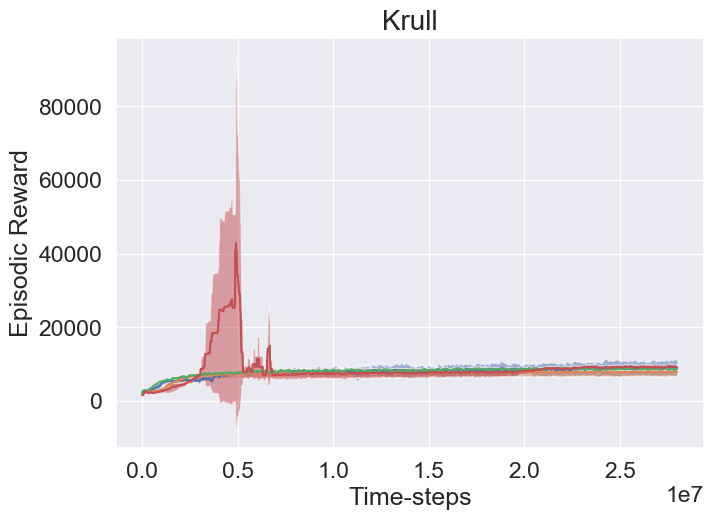}
\label{fig:Krull}
\end{subfigure}%
~
\begin{subfigure}[t]{0.165\textwidth}
\centering
\includegraphics[width=\textwidth]{KungFuMaster}
\label{fig:KungFuMaster}
\end{subfigure}%
~
\begin{subfigure}[t]{0.165\textwidth}
\centering
\includegraphics[width=\textwidth]{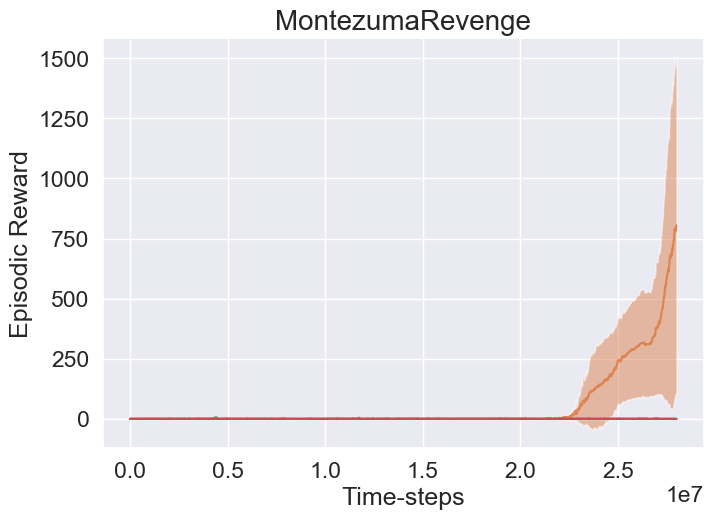}
\label{fig:MontezumaRevenge}
\end{subfigure}%
~
\begin{subfigure}[t]{0.165\textwidth}
\centering
\includegraphics[width=\textwidth]{MsPacman}
\label{fig:MsPacman}
\end{subfigure}%

\begin{subfigure}[t]{0.165\textwidth}
\centering
\includegraphics[width=\textwidth]{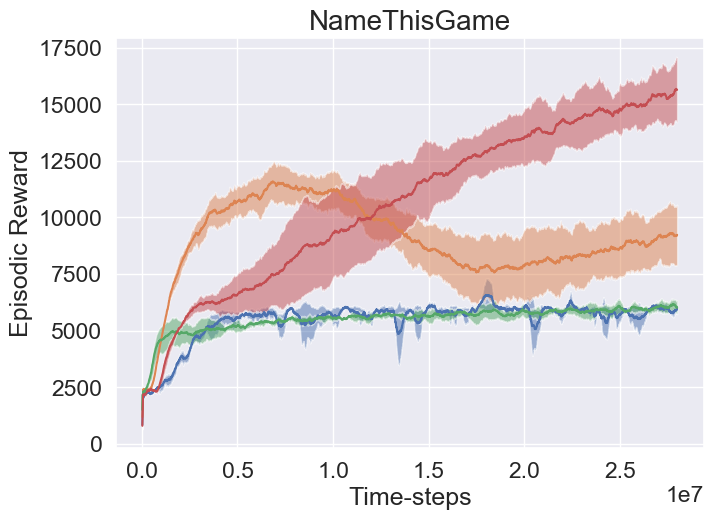}
\label{fig:NameThisGame}
\end{subfigure}%
~
\begin{subfigure}[t]{0.165\textwidth}
\centering
\includegraphics[width=\textwidth]{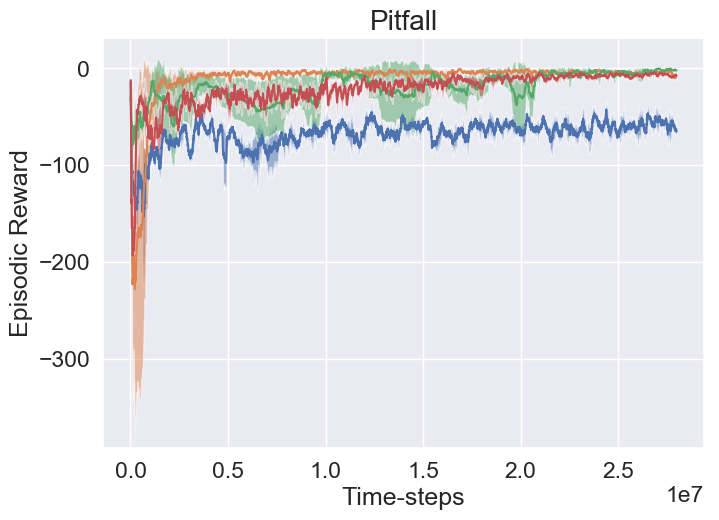}
\label{fig:Pitfall}
\end{subfigure}%
~
\begin{subfigure}[t]{0.165\textwidth}
\centering
\includegraphics[width=\textwidth]{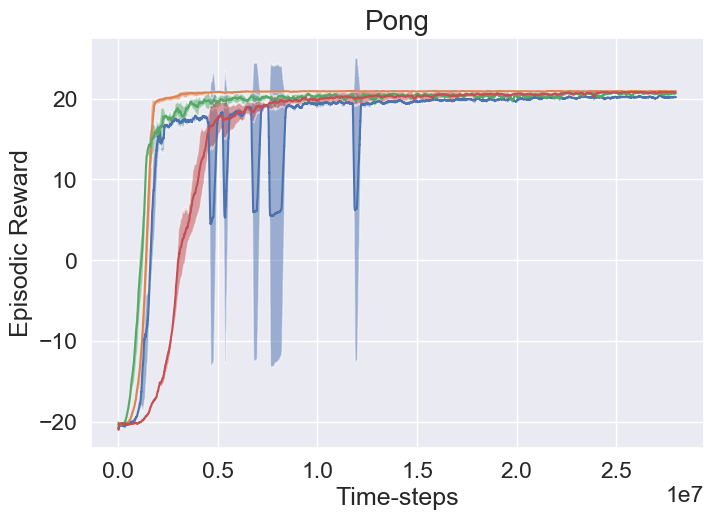}
\label{fig:Pong}
\end{subfigure}%
~
\begin{subfigure}[t]{0.165\textwidth}
\centering
\includegraphics[width=\textwidth]{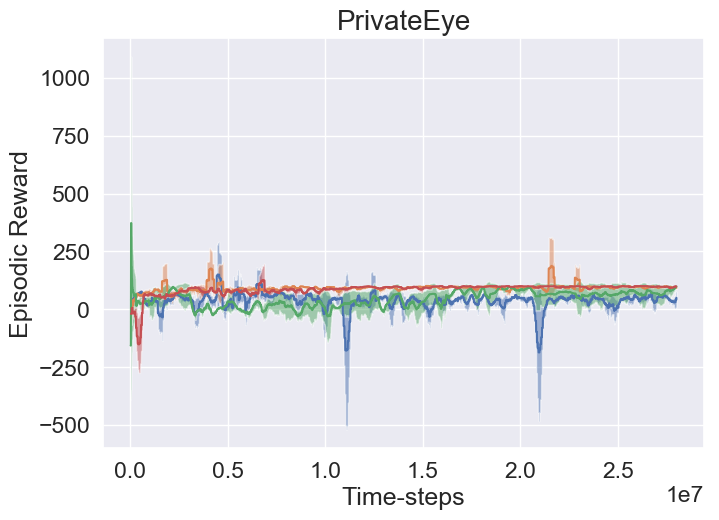}
\label{fig:PrivateEye}
\end{subfigure}
~
\begin{subfigure}[t]{0.165\textwidth}
\centering
\includegraphics[width=\textwidth]{Qbert}
\label{fig:Qbert}
\end{subfigure}

\begin{subfigure}[t]{0.165\textwidth}
\centering
\includegraphics[width=\textwidth]{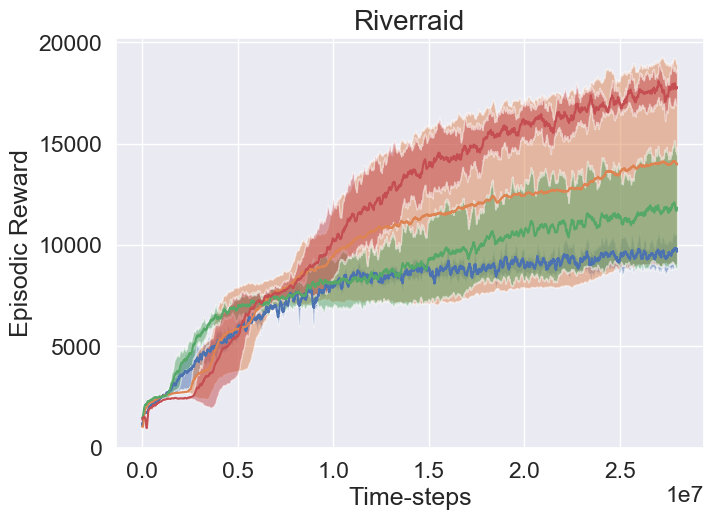}
\label{fig:Riverraid}
\end{subfigure}
~
\begin{subfigure}[t]{0.165\textwidth}
\centering
\includegraphics[width=\textwidth]{RoadRunner}
\label{fig:RoadRunner}
\end{subfigure}
~
\begin{subfigure}[t]{0.165\textwidth}
\centering
\includegraphics[width=\textwidth]{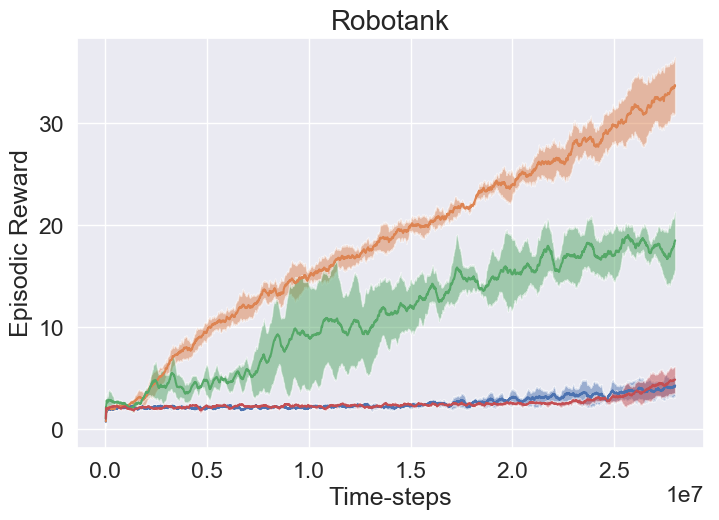}
\label{fig:Robotank}
\end{subfigure}
~
\begin{subfigure}[t]{0.165\textwidth}
\centering
\includegraphics[width=\textwidth]{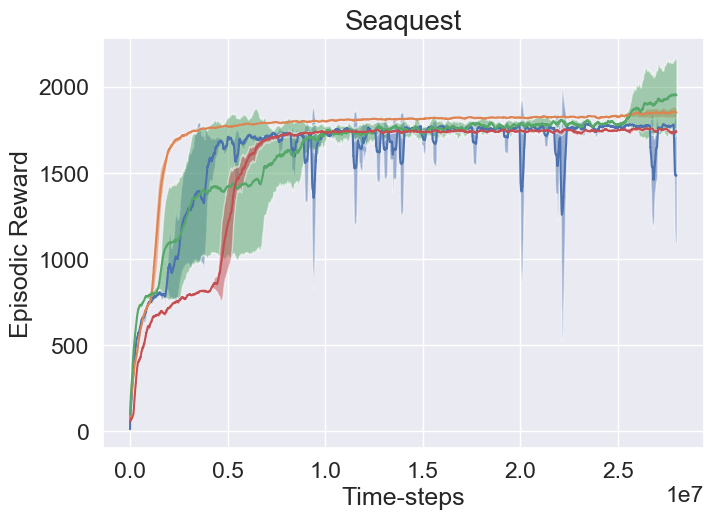}
\label{fig:Seaquest}
\end{subfigure}
~
\begin{subfigure}[t]{0.165\textwidth}
\centering
\includegraphics[width=\textwidth]{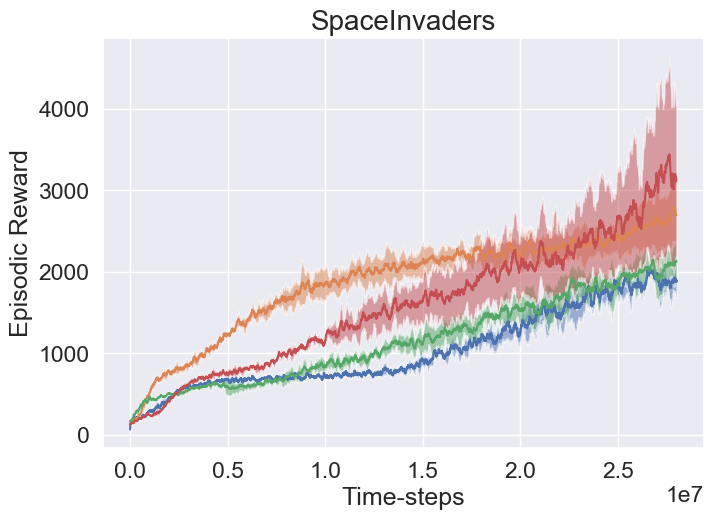}
\label{fig:SpaceInvaders}
\end{subfigure}

\begin{subfigure}[t]{0.165\textwidth}
\centering
\includegraphics[width=\textwidth]{StarGunner}
\label{fig:StarGunner}
\end{subfigure}
~
\begin{subfigure}[t]{0.165\textwidth}
\centering
\includegraphics[width=\textwidth]{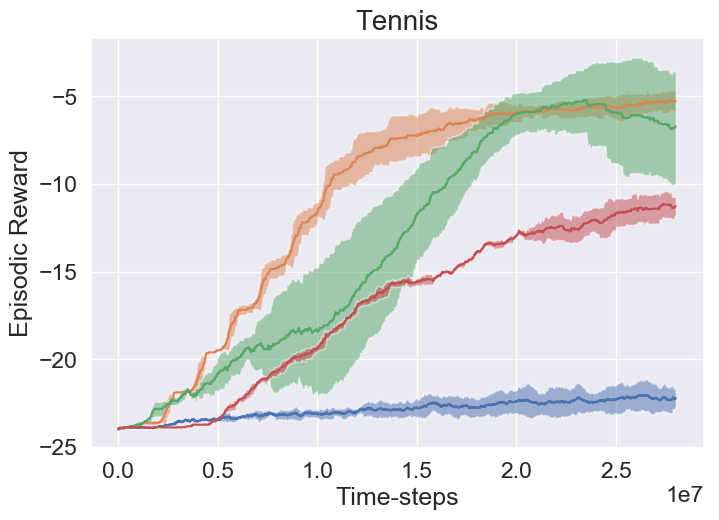}
\label{fig:Tennis}
\end{subfigure}
~
\begin{subfigure}[t]{0.165\textwidth}
\centering
\includegraphics[width=\textwidth]{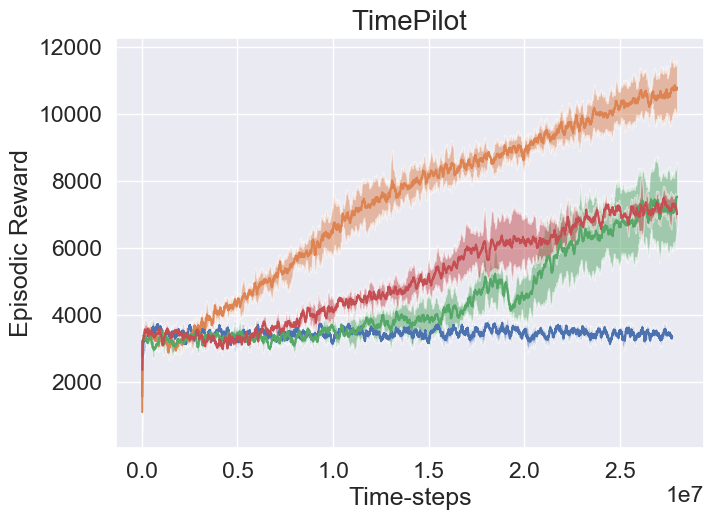}
\label{fig:TimePilot}
\end{subfigure}
~
\begin{subfigure}[t]{0.165\textwidth}
\centering
\includegraphics[width=\textwidth]{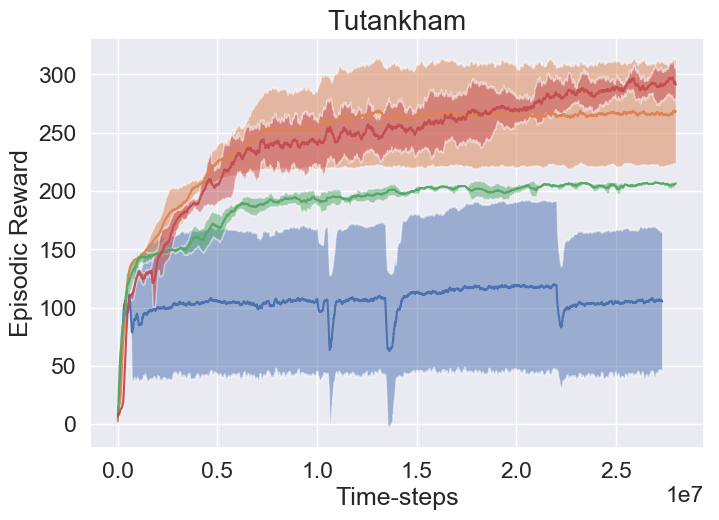}
\label{fig:Tutankham}
\end{subfigure}
~
\begin{subfigure}[t]{0.165\textwidth}
\centering
\includegraphics[width=\textwidth]{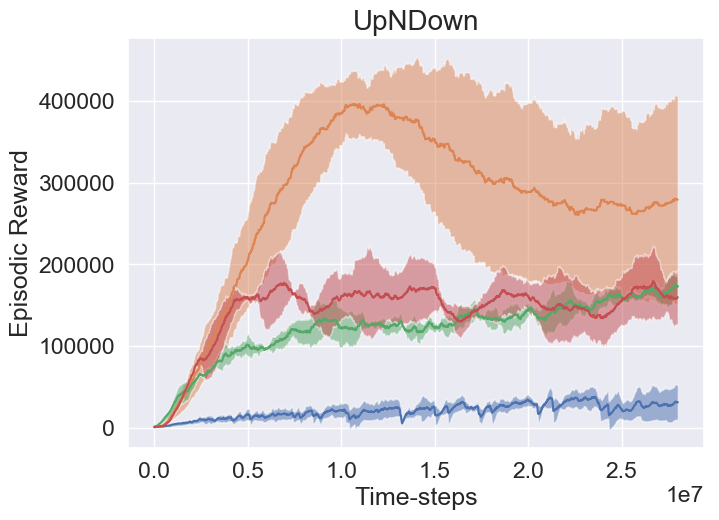}
\label{fig:UpNDown}
\end{subfigure}

\begin{subfigure}[t]{0.165\textwidth}
\centering
\includegraphics[width=\textwidth]{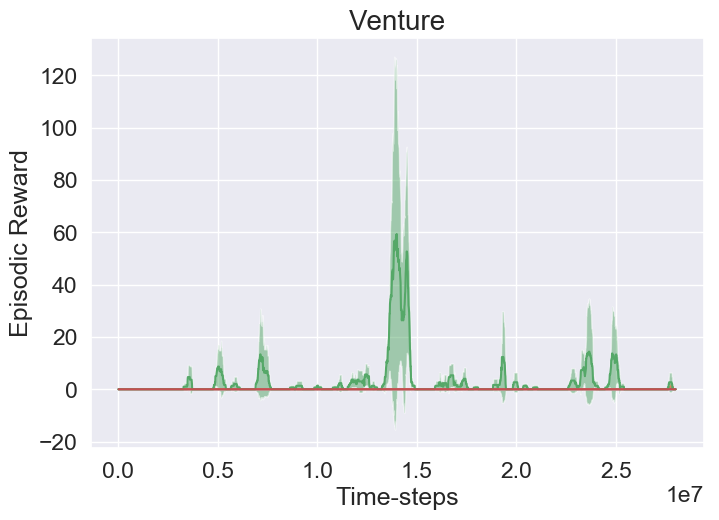}
\label{fig:Venture}
\end{subfigure}
~
\begin{subfigure}[t]{0.165\textwidth}
\centering
\includegraphics[width=\textwidth]{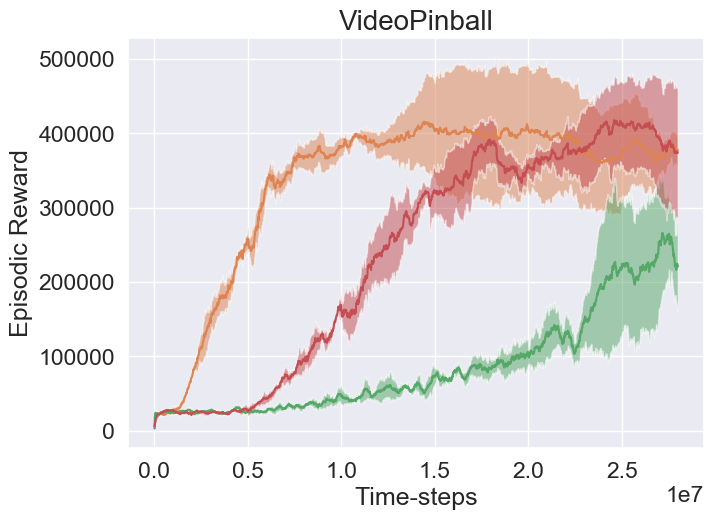}
\label{fig:VideoPinball}
\end{subfigure}
~
\begin{subfigure}[t]{0.165\textwidth}
\centering
\includegraphics[width=\textwidth]{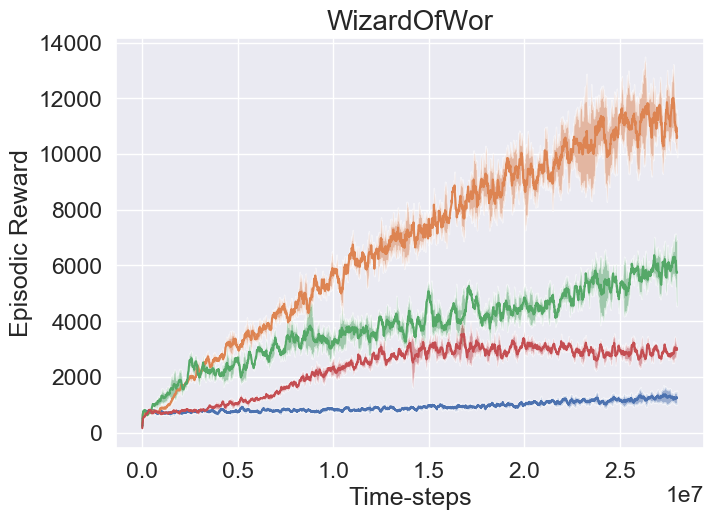}
\label{fig:WizardOfWor}
\end{subfigure}
~
\begin{subfigure}[t]{0.165\textwidth}
\centering
\includegraphics[width=\textwidth]{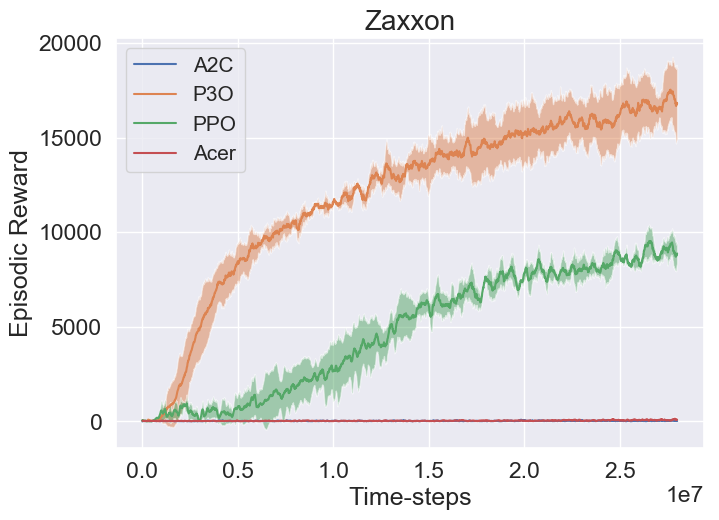}
\label{fig:Zaxxon}
\end{subfigure}
~
\caption{\tbf{Training curves of A2C (blue), ACER (red), PPO (green) and P3O (orange) on all 49 Atari games.}}
\label{fig:atari}
\end{figure*}
\end{appendix}

\end{document}